\documentclass[sigconf, nonacm]{acmart}

\usepackage{makecell}
\usepackage{multirow}
\usepackage{colortbl}
\usepackage{amsmath,amsfonts}
\usepackage{textcomp}
\usepackage{xspace}
\usepackage{stfloats}
\usepackage{url}
\usepackage{booktabs}
\usepackage{multirow}
\usepackage{color}
\usepackage{xcolor}
\usepackage{verbatim}
\usepackage{graphicx}
\usepackage{algorithm}
\usepackage{algpseudocode}
\usepackage{enumitem}
\usepackage{subcaption}
\usepackage{array}
\usepackage{hyperref}
\definecolor{MyRed}{HTML}{a22f2c}
\definecolor{MyRed2}{HTML}{e22046}
\definecolor{MyYellow}{HTML}{9e7132}
\definecolor{MyGreen}{HTML}{387f61}
\definecolor{MyBlue}{HTML}{4D65EF} 
\definecolor{fbApp}{HTML}{ffe4e3}
\definecolor{tabhighlight}{HTML}{e5e5e5}
\newcommand\vldbdoi{XX.XX/XXX.XX}
\newcommand\vldbpages{X-X}
\newcommand\vldbvolume{14}
\newcommand\vldbissue{1}
\newcommand\vldbyear{2020}
\newcommand\vldbauthors{\authors}
\newcommand\vldbtitle{\shorttitle} 
\newcommand\vldbavailabilityurl{URL_TO_YOUR_ARTIFACTS}

\newcommand{\splitcell}[2]{%
  \begin{tabular}{@{}c@{}}#1\\ \cmidrule(lr){1-1} #2\end{tabular}%
}
\newcommand\vldbpagestyle{plain} 

\newcommand{\ie}{\textit{i.e.,}\xspace}
\newcommand{\eg}{\textit{e.g.,}\xspace}
\newcommand{\rowc}{\rowcolor{fbApp}}

\newcommand{\mypara}[1]{\noindent\textbf{#1} }

\begin{document}
\title{QUALS: Corpus Equilibrium for Universal Forecasting via Pattern Quantization and Learnability Synchronization}

\author{Yujie Li$^{1,2}$, Zezhi Shao$^{1\ast}$, Chengqing Yu$^{1,2}$, Yisong Fu$^{1,2}$, Weijie Zhu$^{1}$, Yifan Du$^{1}$, Jilin Hu$^{3}$, Bin Yang$^{3}$, Yongjun Xu$^{1,2}$, Fei Wang$^{1,2\ast}$}
\affiliation{%
  \institution{$^{1}$State Key Laboratory of AI Safety, Institute of Computing Technology, Chinese Academy of Sciences, Beijing, China\\
  $^{2}$University of Chinese Academy of Sciences, Beijing, China\\
  $^3$School of Data Science and Engineering, East China Normal University, Shanghai, China
  }
  \institution{\{liyujie23s, shaozezhi, yuchengqing22b, fuyisong24s, zhuweijie26s, duyifan, xyj, wangfei\}@ict.ac.cn, \\ \{jlhu, byang\}@dase.ecnu.edu.cn}
}
\begin{abstract}
Ubiquitous time series data across diverse domains enables critical applications in areas such as transportation systems and power grids. Recently, training foundation models on massive datasets to achieve accurate zero-shot forecasting has emerged as a major research focus. However, current studies predominantly prioritize architectural innovations while insufficiently addressing data diversity, often relying on simple data sampling strategies that fail to manage complex data distributions effectively, 
leading to inefficient use of training data and suboptimal performance.
%
%
To address this, we propose QUALS, a large-scale time series corpus equilibrium framework.
QUALS significantly enhances data efficiency, \ie enabling existing models to achieve superior performance using only a small fraction of the original training data.
Specifically, QUALS operates through two core mechanisms.
First, an pattern quantization framework systematically decodes heterogeneous patterns from mixed corpora via vector quantization and uniform binning.
Second, a learnability synchronization framework calibrates sampling weights for heterogeneous patterns, which bridges the optimization gap between simple and complex motifs to maximize overall training efficiency.
Extensive benchmarks demonstrate that pre-training on QUALS consistently achieves superior zero-shot performance, even under substantially reduced training budgets.
\end{abstract}

\renewcommand*{\authors}{Yujie Li, Zezhi Shao. Chengqing Yu, Yisong Fu, Weijie Zhu, Yifan Du, Jilin Hu, Bin Yang, Yongjun Xu, Fei Wang} 
\maketitle

\pagestyle{\vldbpagestyle}
\begingroup\small\noindent\raggedright\textbf{PVLDB Reference Format:}\\
\vldbauthors. \vldbtitle. PVLDB, \vldbvolume(\vldbissue): \vldbpages, \vldbyear.\\
\href{https://doi.org/\vldbdoi}{doi:\vldbdoi}
\endgroup
\begingroup
\renewcommand\thefootnote{}\footnote{\noindent
$^*$: Corresponding author. \\
This work is licensed under the Creative Commons BY-NC-ND 4.0 International License. Visit \url{https://creativecommons.org/licenses/by-nc-nd/4.0/} to view a copy of this license. For any use beyond those covered by this license, obtain permission by emailing \href{mailto:info@vldb.org}{info@vldb.org}. Copyright is held by the owner/author(s). Publication rights licensed to the VLDB Endowment. \\
\raggedright Proceedings of the VLDB Endowment, Vol. \vldbvolume, No. \vldbissue\ %
ISSN 2150-8097. \\
\href{https://doi.org/\vldbdoi}{doi:\vldbdoi} \\
}\addtocounter{footnote}{-1}\endgroup

\ifdefempty{\vldbavailabilityurl}{}{
\vspace{.3cm}
\begingroup\small\noindent\raggedright\textbf{PVLDB Artifact Availability:}\\
Source code: \url{https://github.com/blisky-li/QUALS}\\
\endgroup
}

\section{Introduction}
Time series data are fundamental to a wide range of critical real-world systems, including transportation management~\cite{fang2021mdtp, shao2022decoupled}, cloud resource monitoring~\cite{poppe2020seagull, xie2025chatts}, public safety~\cite{zheng2023declog, jensen2025demonstration}, and beyond~\cite{Travel,Demand,miao2024less}. The strong heterogeneity~\cite{shao2024exploring, qiu2024tfb} of these applications has made universal time series forecasting\footnote{While these models are referred to as foundational or general models in some prior work, we adopt the term universal forecasting models~\cite{gifteval, UniTS, ROSE, shao2025blast, woo2024moirai} throughout this paper for consistency and to avoid confusion with multi-task models.
} an increasingly important goal: rather than building specialized models for each domain~\cite{li2025apt, li2025ufgtime, wu2023autocts+}, we seek models that can generalize across heterogeneous domains and deliver strong zero-shot forecasting~\cite{ansari2025chronos2, liu2024moiraimoe} performance on previously unseen data.

\begin{figure}[t]
  \centering
  
  \includegraphics[width=1\linewidth]{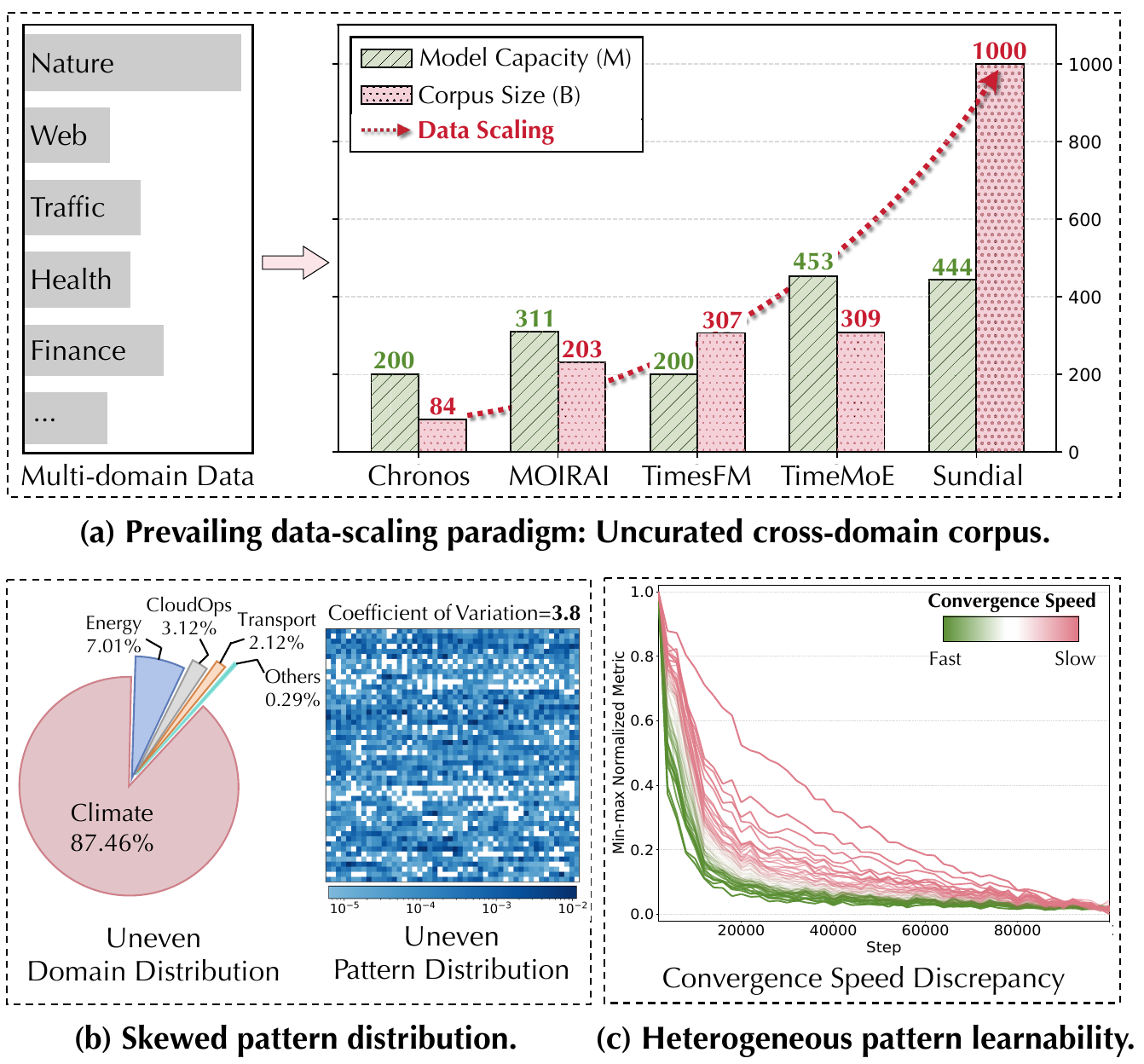}
  \caption{
  Overview of the data-scaling paradigm, highlighting skewed pattern distributions and imbalanced pattern learnability. In (b), domain shares are derived from GIFT-Eval Pretrain metadata, while the pattern distribution is measured using the analysis introduced later and projected into 2-D.
  }
  \label{fig:intro}
\end{figure}

Recent progress has largely followed a foundation-model paradigm: train large models over massive corpora and rely on scale to induce emergent zero-shot capability. Representative models such as Chronos~\cite{Chronos}, MOIRAI~\cite{woo2024moirai}, and Sundial~\cite{liu2025sundial} all perform large-scale pre-training based on different Transformer architectures and have achieved state-of-the-art performance.
%
Notably, unlike natural language processing~\cite{hoffmann2022training}, where progress is often associated with scaling model capacity, universal time series forecasting has seen a more pronounced trend toward expanding the \textit{training corpus}.
To cope with the heterogeneity of cross-domain data and improve zero-shot performance, existing studies tend to typically aggregate open-source datasets from diverse domains and continuously expand corpus size~\cite{TimeGPT1, TimesFM, sun2025xihe, shi2024timemoe}, as illustrated in Figure~\ref{fig:intro}(a). For example, the above three models contain only 200M, 311M, and 444M parameters, respectively, while their corresponding training corpora reach 84B, 231B, and even 1032B points.

This data scale-driven strategy has improved performance, but it also reveals a fundamental limitation: the training distribution is not guaranteed to be well aligned with the needs of universal forecasting, while effectively leveraging larger corpora typically requires substantially greater computational cost.
In particular, current corpora suffer from two coupled forms of imbalance.

First, \textit{pattern distribution is highly skewed}~\cite{shao2025blast}. Expanding source domains does not necessarily increase the diversity of underlying temporal patterns. In large aggregated corpora, a few dominant motifs can occupy most of the probability mass, resulting in a pronounced long-tail distribution. 
%
As shown in Figure 1(b), climate data account for 87\% of the GIFT-Eval Pretrain corpus~\cite{gifteval}. This source-level imbalance is also reflected at the pattern level, where the measured distribution has a coefficient of variation (COV) of 3.8, indicating strong non-uniformity.
Under standard sampling~\cite{woo2024moirai}, models are repeatedly exposed to frequent patterns, while rare but important ones remain underrepresented, ultimately limiting zero-shot generalization.
Second, \textit{heterogeneous patterns differ substantially in learnability}~\cite{wang2025aries}. Time series vary widely in complexity: regular motifs are typically easy to fit, whereas patterns with multiple seasonalities or non-stationarity are much harder to optimize~\cite{wang2025fredf, fu2025selective}. Thus, even with a distributionally balanced corpus, easy patterns may dominate training while difficult ones remain under-trained. As shown in Figure~\ref{fig:intro}(c), Chronos exhibits markedly different convergence speeds across pattern types (each line),
%
%
%
revealing a learnability gap between simple and complex motifs.

These observations suggest that the key issue is not corpus size alone, but how to \textit{manage data diversity} for pre-training. Unfortunately, existing corpus construction strategies rarely address this problem in a principled way.
Simple strategies, including naive sampling~\cite{MOMENT,shi2024timemoe} and stratified sampling~\cite{Chronos, woo2024moirai}, fail to address pattern-level imbalance, while recent heuristic balancing methods~\cite{shao2025blast} depend on handcrafted statistical descriptors that are costly to compute and often inadequate for capturing deep temporal semantics.
More importantly, they do not account for learnability differences across patterns.

In this paper, we propose QUALS, a pattern quantization and learnability synchronization corpus equilibrium framework for universal forecasting that explicitly targets both sources of imbalance. QUALS is built on the premise that an efficient pre-training corpus should be balanced not only in terms of what patterns it contains, but also in terms of how those patterns are learned. To achieve this, QUALS introduces two complementary mechanisms. 
First, QUALS performs data-driven \textit{pattern quantization and binning}, which maps highly heterogeneous time series into a discrete pattern space and partitions this space into balanced pattern bins through a multi-scale quantization architecture~\cite{DBLP:conf/aistats/LeeMA23, DBLP:conf/nips/OordVK17, DBLP:conf/iccv/LiuL00W0LG21} coupled with a spectrum-aware strategy~\cite{hotelling1933analysis, DBLP:journals/pami/JegouDS11}. This design replaces expensive heuristic characterization with a scalable, representation-driven alternative, while mitigating the long-tail skew in the raw corpus. Second, QUALS introduces \textit{learnability synchronization}, which profiles the convergence behavior of individual pattern bins during proxy training and adaptively adjusts sampling weights to prioritize difficult patterns while downweighting overly easy ones, based on an offline Group Relative Policy Optimization~(GRPO) strategy~\cite{fu2025reward}. In this way, QUALS aligns learning progress across the corpus and reduces redundant optimization on already saturated patterns.

By jointly balancing pattern distribution and learnability, QUALS substantially improves the data efficiency of universal forecasting. Across extensive benchmarks, models pre-trained on QUALS consistently achieve stronger and more stable zero-shot performance under reduced training budgets; in some settings, it enables superior performance using only 10\% of the original training data.

The main contributions of this paper are as follows:
\begin{itemize}
    \item We identify a previously underexplored \textit{dual imbalance} in universal forecasting corpora—skewed pattern distributions and heterogeneous learnability across patterns—and propose QUALS, a new framework for efficient cross-domain corpus management.
    \item We develop a scalable, data-driven \textit{pattern quantization} method that replaces expensive handcrafted heuristics with a more efficient alternative, effectively mitigating the long-tail issue in large-scale time series data.
    \item We introduce a \textit{learnability synchronization} mechanism that adaptively aligns optimization difficulty across heterogeneous patterns, thereby improving training efficiency under fixed computational budgets.
    \item Extensive experiments show that QUALS consistently improves the data efficiency of universal time series models, delivering more accurate zero-shot forecasting performance while substantially reducing training cost.
\end{itemize}


\section{Related Work}
\label{sec:rw}
\subsection{Universal Time Series Forecasting}
Universal time series forecasting~\cite{li2025tsfm} aims to build foundation models with robust zero-shot generalization across unseen domains. The fundamental paradigm relies on training advanced neural architectures over massive datasets to induce these zero-shot capabilities.

Mainstream models predominantly adopt Transformer architectures across three paradigms: \textit{encoder-only} models learning transferable representations via masked modeling or reconstruction  \cite{MOMENT,liu2024moiraimoe,woo2024moirai, wang2025towards}; \textit{decoder-only} models treating forecasting as conditional autoregressive generation \cite{TimesFM,shi2024timemoe,liu2024timer,DBLP:journals/corr/abs-2511-11698,TimeXL}; and \textit{encoder-decoder} models using series discretization to frame probabilistic forecasting as sequence-to-sequence tasks  \cite{Chronos, ansari2025chronos2, DBLP:conf/icml/0004Q00RP0G25}. Complementary works explore alternative inductive biases, including prior-based meta-forecasting and cross-modal transfer  \cite{DAM,ForecastPFN,TTM,DBLP:conf/icml/ChenS0WSL25}

Despite these architectural divergences, the consensus is that the emergence of zero-shot capabilities is fundamentally driven by data rather than mere structural design. Consequently, a prevalent trend in current research is to blindly scale up training pipelines by aggregating massive, multi-domain corpora to enhance performance. However, this unchecked data scaling exposes critical vulnerabilities in data distribution and utilization efficiency, necessitating a closer examination of pre-training corpora

\subsection{Time Series Pre-training Corpora}
Time-series corpora underpin universal forecasting capability. Most pipelines follow a scale-driven paradigm---LOTSA for MOIRAI~\cite{woo2024moirai}, the Chronos corpus~\cite{Chronos}, Time-300B for TimeMoE~\cite{shi2024timemoe}, and UTSD for Timer~\cite{liu2024timer}---yet source-proportional or naive exposure still over-represents repetitive motifs and wastes pre-training budget. Subsequent work refines balancing at progressively finer granularity: dataset level (capped LOTSA stratification in MOIRAI and dataset-uniform training in Chronos~\cite{woo2024moirai,Chronos}), domain level (UTSD in Timer~\cite{liu2024timer}), and pattern level via statistical descriptors (BLAST~\cite{shao2025blast} and catch22~\cite{lubba2019catch22}). These strategies reduce redundancy but remain limited: dataset-/domain-level control is still coarse-grained and dominant within-domain motifs may persist; heuristic pattern bucketing is costly at billion scale and may miss deep temporal semantics~\cite{shao2025blast}. They also largely ignore \textit{learnability discrepancy}---easy motifs converge faster than hard ones~\cite{cheng2023weakly,wang2025aries,qiu2025easytime}---so uniform exposure leaves difficult patterns under-optimized.

\begin{table}[t]
  \centering
  \setlength{\abovecaptionskip}{0.cm}
  \setlength{\belowcaptionskip}{0.cm}
    \renewcommand\arraystretch{1}
  \caption{Comparison of corpus curation paradigms. ``Others'' denotes scale-driven corpora with naive sampling or dataset-/domain-level balancing~\cite{Chronos,woo2024moirai,shi2024timemoe,liu2024timer}.}
\label{tab:comparison}
\resizebox{\linewidth}{!}{
\begin{tabular}{c|c|c|c}
\toprule[1.2pt]
\textbf{Challenges} & \begin{tabular}[c]{@{}c@{}}\textbf{QUALS}\\ \textbf{(Ours)}\end{tabular} & \begin{tabular}[c]{@{}c@{}}\textbf{BLAST}\\ 
{~\cite{shao2025blast}}\end{tabular} & \begin{tabular}[c]{@{}c@{}}\textbf{Others}\\ {\cite{Chronos,woo2024moirai,shi2024timemoe,liu2024timer}}\end{tabular} \\
\midrule
\begin{tabular}[c]{@{}c@{}}Distribution\\ Imbalance\end{tabular}& Deep Data-Driven  & Heuristic & Dataset/Domain-level \\
\midrule
\begin{tabular}[c]{@{}c@{}}Learnability\\ Synchronization\end{tabular} & GRPO-Synchronized  & $\times$ & $\times$ \\
\bottomrule[1.2pt]
\end{tabular}
}

\end{table}

Table~\ref{tab:comparison} summarizes these gaps: prior paradigms remain scale-centric and do not jointly address pattern distribution and learnability. QUALS targets both via data-driven pattern quantization and GRPO-synchronized learnability-aware curation.

\section{Preliminaries}
\label{sec:preliminaries}
\subsection{Universal Time Series Forecasting}
Let $\mathcal{D}_{train}$ denote a large-scale pre-training distribution of time series spanning diverse domains. For an instance $x \in \mathbb{R}^{L}$, universal forecasting aims to learn a global function $f_\theta$, parameterized by model weights $\theta$, that predicts future values $x_{future} \in \mathbb{R}^{H}$ given the historical context $x_{history} \in \mathbb{R}^{C}$.

The training objective minimizes the expected risk over the data distribution defined by a loss function $\mathcal{L}$:
\begin{equation}
    \min_{\theta} \mathbb{E}_{x \sim \mathcal{D}_{train}} [\mathcal{L}(f_\theta(x_{history}), x_{future})].
\end{equation}

Unlike domain-specific forecasting~\cite{li2024dynamic,shao2024exploring}, $f_\theta$ must make accurate forecasts in a zero-shot manner.

\begin{table}[t]
    \centering
      \setlength{\abovecaptionskip}{0.cm}
      \setlength{\belowcaptionskip}{0.cm}
    \renewcommand\arraystretch{0.9}
    \caption{Key notations in QUALS.}
    \label{notation}
    \setlength{\tabcolsep}{3pt}
    \resizebox{\columnwidth}{!}{
    \begin{tabular}{c|c|l}
        \toprule[1.2pt]
         \multicolumn{2}{c|}{\textbf{Notation}} & \multicolumn{1}{c}{\textbf{Description}} \\
        \midrule

        \multirow{8}{*}{\rotatebox{90}{Stage 1}}
        & $x$ & Input time series instance \\
        & $L$ & Length of time series input \\
        & $P, S$ & Patch size and stride size \\
        & $H$ & Number of hierarchical stages in Swin Encoder \\
        & $D$ & Hidden layer dimension of Swin \\
        & $C, \tilde{C}$ & Learnable codebook and its spectrally rotated version \\
        & $K, d$ & Codebook size and embedding dimension ($d \ll D$) \\
        & $\mathbf{Z}_{ms}$ & Multi-scale patch representations extracted by Swin \\
        \midrule

        \multirow{6}{*}{\rotatebox{90}{Stage 2}}
        & $\Sigma$ & Empirical covariance matrix of the codebook \\
        & $\lambda_i, \mathbf{v}_i$ & $i$-th eigenvalue and eigenvector of $\Sigma$ \\
        & $\mathbf{b}$ & Vector of bin counts allocated to each latent dimension \\
        & $\mathcal{E}$ & Set of ECDF-based partition boundaries \\
        & $\mathcal{U}$ & Final set of Pattern Bins, $\mathcal{U} = \{u_1, \dots, u_M\}$ \\
        & $M$ & Target number of pattern bins (Capacity constraint) \\
        \midrule

        \multirow{6}{*}{\rotatebox{90}{Stage 3}}
        & $v_i$ & Static embedding of bin $u_i$ for policy network \\
        & $\pi_\theta$ & Policy network for dynamic weight adjustment \\
        & $w_i$ & Sampling weight of bin $u_i$ generated by policy $\pi_\theta$ \\
        & $r_{i,t}$ & Convergence speed of bin $u_i$ at iteration $t$ \\
        & $R_{i,t}$ & Reward signal derived from relative learning speed \\
        & $\mathcal{H}$ & Group Pool storing historical experiences for GRPO \\
        \bottomrule[1.2pt]
    \end{tabular}
    }
\end{table}

\subsection{The Corpus Equilibrium Problem}

Raw aggregated corpora $\mathcal{D}_{raw}$ suffer from \textit{pattern imbalance}, where frequent motifs dominate gradients, and \textit{learnability discrepancy}, where simple patterns converge significantly faster.

We formulate \textit{corpus equilibrium} as a two-stage policy optimization problem. The first policy, denoted by $\pi_{c}$, transforms the raw corpus $\mathcal{D}_{raw}$ into a pattern-balanced candidate corpus $\mathcal{D}_{cand}$. The second policy, denoted by $\pi_{s}$, assigns sampling weights $w_i \in [0,1]$ to pattern bins in $\mathcal{D}_{cand}$ to control the actual pre-training subset selection. Accordingly, the objective is to jointly find a corpus construction policy and a sampling policy that minimize the zero-shot prediction error on the validation benchmark $\mathcal{D}_{val}$ under a strict training budget $B$:
\begin{equation}
\min_{\pi_c,\pi_s}
\;
\mathbb{E}_{\substack{\mathcal{D}_{cand} \sim \pi_c(\mathcal{D}_{raw})\\
S \sim \pi_s(\mathcal{D}_{cand})}}
\![\mathcal{E}(\mathcal{D}_{val}; f_{\theta_S})]
\quad
\textit{s.t.}
\quad |S| \le B ,
\end{equation}
where $\mathcal{D}_{cand}$ is the candidate corpus constructed from $\mathcal{D}_{raw}$, $S$ is the subset sampled from $\mathcal{D}_{cand}$, $\theta_S$ denotes the model parameters optimized on $S$, and $\mathcal{E}(\mathcal{D}_{val}; f_{\theta_S})$ is the zero-shot validation error. In QUALS, $\pi_c$ is instantiated by \textit{pattern quantization}, while $\pi_s$ is optimized by \textit{learnability synchronization} using pattern-specific convergence speeds $r_{i,t}$.
This objective separates corpus equilibrium into pattern equilibrium ($\pi_c$: balanced bins in $\mathcal{D}_{cand}$) and learnability equilibrium ($\pi_s$: budget allocation under $|S|\le B$).
Ideally, the learnability of bin $u_i$ is the minimum budget to reach a validation threshold,
\[
B_i^\star=\min\{B:\ell_i(B)\le \tau_i\},
\]
where $\ell_i(B)$ is the validation loss on $u_i$ after budget $B$ and $\tau_i$ is a convergence threshold.
At billion scale, $B_i^\star$ is not directly observable, so we use the uniform-probing convergence speed
\[
r_{i,t}=-\frac{\Delta \ell_i(t)}{\Delta t}
\]
(larger $r_{i,t}$: faster under equal exposure). Policy $\pi_s$ then sets $w_i$ to synchronize the weighted speeds $w_i r_{i,t}$.
Key mathematical notations used throughout this paper are summarized in Table~\ref{notation}.

\begin{figure*}[t]
  \setlength{\abovecaptionskip}{0.2cm}
  \setlength{\belowcaptionskip}{0.cm}
  \centering
  \includegraphics[width=1\linewidth]{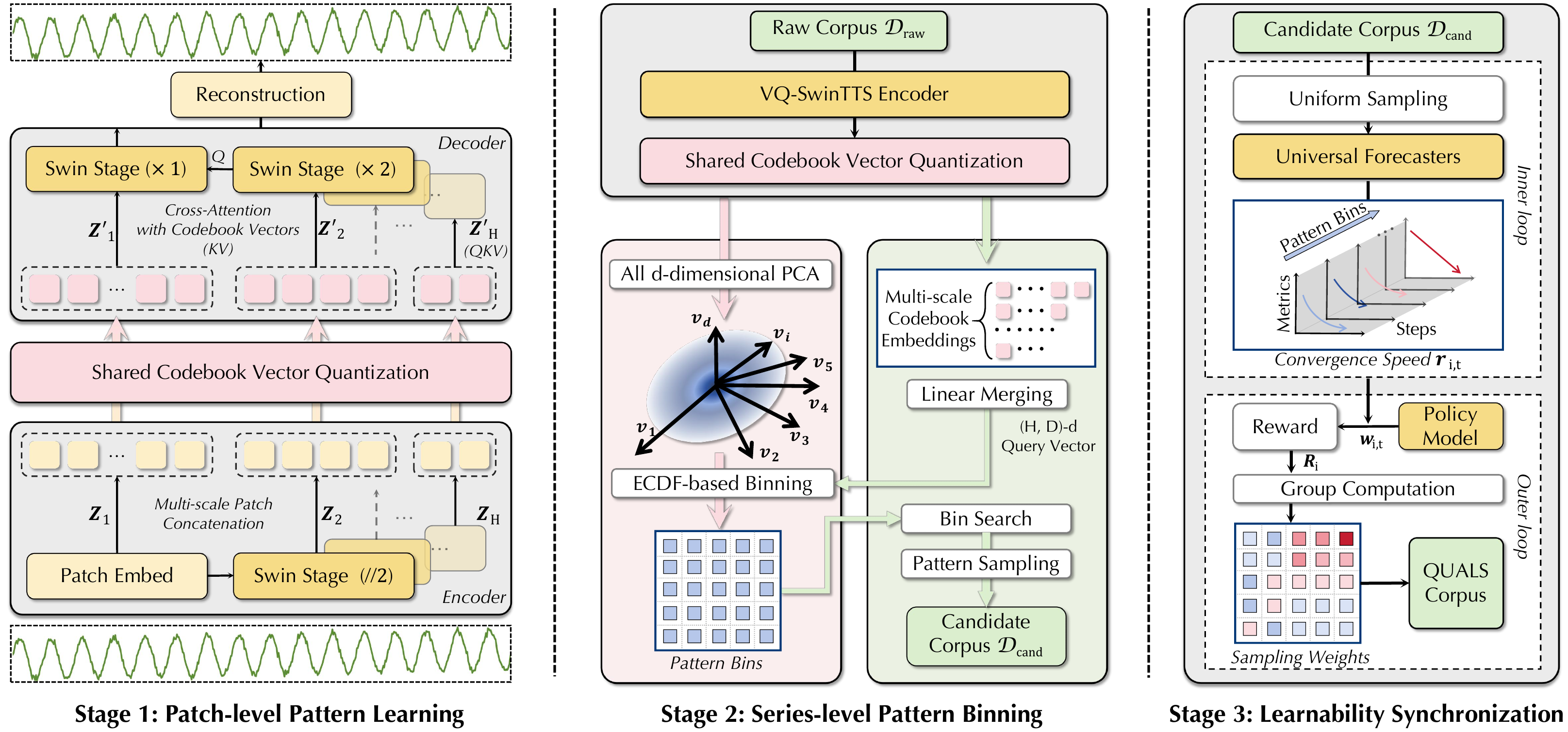}
  \caption{Overview of the QUALS framework, a highly scalable, GPU-accelerated pipeline for billion-scale time series pre-training corpus construction. It systematically integrates deep temporal quantization via VQ-SwinTTS (Stage 1), spectral pattern binning (Stage 2), and GRPO-based learnability synchronization (Stage 3) to construct a pattern-balanced and learnability-synchronized corpus for efficient universal forecasting pre-training.}
  \vspace{-5pt}
  \label{fig:main}
\end{figure*}

\section{Method}
\label{sec:method}
\subsection{Overview}
QUALS constructs a pre-training corpus with a balanced pattern distribution from large-scale time series to improve zero-shot generalization. We divide this process into three progressive stages: pattern representation, pattern sampling, and pattern rebalancing.
As illustrated in Figure~\ref{fig:main}, QUALS achieves this goal through:

\noindent\textbf{(1) Patch-Level Pattern Discretization.}
We transform continuous time series into discrete codes via \textit{VQ-SwinTTS} (Figure~\ref{fig:main}, left), which uses a hierarchical Swin encoder to extract multi-scale patch embeddings and quantizes them with a shared codebook.

\noindent\textbf{(2) Series-Level Pattern Binning.}
We conduct globally uniform pattern space by projecting the learned codebook onto PCA components (Figure~\ref{fig:main}, middle), and assign it into ECDF-defined semantic bins to obtain density-balanced pattern partitions.

\noindent\textbf{(3) Learnability Synchronization.}
We address optimization imbalance across bins with \textit{GRPO-based Synchronization} (Figure~\ref{fig:main}, right): a policy network adjusts bin sampling weights using convergence speeds as rewards, which allocates more focus to challenging patterns and synchronizes learning progress across the corpus.

This three-stage structure follows the two-policy formulation in Section~\ref{sec:preliminaries}: Stages~1--2 implement $\pi_c$ through pattern bins, and Stage~3 implements $\pi_s$ through learnability-aware bin weights.
\subsection{VQ-SwinTTS}~\label{vqswintts}

Effective pattern balancing depends on a representation method that captures temporal features beyond heuristic properties. To this end, we introduce \textbf{V}ector-\textbf{Q}uantized \textbf{Swin} \textbf{T}ransformer for \textbf{T}ime-\textbf{S}eries~({VQ-SwinTTS}) to learn deep time-series semantics. Its design jointly supports multi-scale dependency modeling within individual series and sufficient pattern differentiation in the learned representation space: Swin captures temporal structures across scales, and vector quantization induces a discrete and collapse-resistant codebook for subsequent balancing.
Stage~1 thus converts raw series into reusable multi-scale discrete codes.

\mypara{Pre-processing.}
Using BLAST corpus as training samples~\cite{shao2025blast}, we pad or truncate variable-length time series~\cite{miao2025spatio} to fixed-length inputs~$x \in \mathbb{R}^L$ for batch processing. These sequences are unfolded into patches~$x_{p} \in \mathbb{R}^{N \times P}$ with stride~$S$~\cite{DBLP:conf/ijcai/CirsteaG0KDP22}, mapped to embeddings~$\mathbf{Z}_{input} \in \mathbb{R}^{N \times D}$ augmented with learnable positional encodings. It should be noted that BLAST is used only for training VQ-SwinTTS, whereas the subsequent corpus-level encoding and balancing are carried out on the full raw corpus.

\mypara{Swin Encoder.}
We use shifted-window self-attention (Swin~\cite{DBLP:conf/iccv/LiuL00W0LG21}) to extract multi-scale patch representations. The encoder has $H$ blocks at progressively coarser temporal resolutions. Each block applies W-MSA and SW-MSA, followed by \textit{patch merging} to halve the temporal resolution and expand channels. Instead of discarding intermediate features, we concatenate all patch embeddings $\mathbf{Z}_{(h)}$ from each block $h\in\{1,\dots,H\}$:
\begin{equation}
\mathbf{Z}_{\mathrm{ms}} = \bigoplus_{h=1}^{H} \mathbf{Z}_{(h)} \in \mathbb{R}^{N_{ms}\times D}, N_{ms} = \sum^{H}_{h=1}N_{h},
\end{equation}
which forms the continuous multi-scale representation to be discretized by the subsequent VQ module.

\mypara{Shared Codebook Vector Quantization.}
We discretize the aggregated multi-scale representations $\mathbf{Z}_{\mathrm{ms}}$ using a single shared codebook~\cite{DBLP:conf/nips/OordVK17}. Formally, we define a learnable codebook $C$:
\begin{equation}
C = [q_1, \dots, q_K] \in \mathbb{R}^{K \times d}.
\end{equation}
For each patch embedding $\mathbf{z}_e \in \mathbb{R}^{D}$ from $\mathbf{Z}_{\mathrm{ms}}$, the quantization layer identifies the optimal codebook index via a nearest-neighbor search based on cosine similarity~\cite{DBLP:conf/iclr/YuLKZPQKXBW22}. To mitigate codebook collapse~\cite{dhariwal2020jukebox} and facilitate pattern space construction, we employ linear transformations $\mathbf{W}_{in} \in \mathbb{R}^{D \times d}$ and $\mathbf{W}_{out} \in \mathbb{R}^{d \times D}$ for dimensionality decoupling. The quantization process is formulated as:

\begin{equation}
\begin{aligned}
k & = \arg\max_{k \in \{1,\dots,K\}}
\frac{
\langle \mathbf{W}_{in}\mathbf{z}_e, q_k \rangle
}{
\|\mathbf{W}_{in}\mathbf{z}_e\|_2 \cdot \|q_k\|_2
} \\ 
& = \arg\max_{k \in \{1,\dots,K\}}
\frac{
\langle \mathbf{W}_{in}\mathbf{z}_e, \delta_kC \rangle
}{
\|\mathbf{W}_{in}\mathbf{z}_e\|_2 \cdot \|\delta_kC\|_2
},
\end{aligned}
\end{equation}
where $\delta_k$ denotes the $k$-th one-hot basis vector. The selected codebook vectors $\delta_{ms} C$ is then projected as $\mathbf{Z^{'}_{ms}}=\mathbf{W}_{out} \delta_{ms} C$, which serves as the output of the decoder. This design ensures that the learned patterns remain shape-invariant, focusing on geometric quality rather than absolute magnitudes.

\mypara{Stabilized Codebook Optimization.}
To support stable pattern separation, broad coverage, and robust utilization for subsequent balancing, we redesign the standard VQ pipeline for temporal pattern learning. We impose a \textit{spherical constraint} via cosine similarity to capture shape-invariant temporal patterns. To optimize the resulting manifold, we replace the standard Straight-Through Estimator~\cite{DBLP:journals/corr/BengioLC13} with the \textit{rotation trick}~\cite{DBLP:conf/iclr/FiftyJDILATR25}, enabling more faithful gradient propagation through rotational displacements. Furthermore, we decouple quantization dimensions through a {low-dimensional linear projection} ($\mathbb{R}^{D} \to \mathbb{R}^{d}$)~\cite{DBLP:conf/iclr/YuLKZPQKXBW22}, and maintain codebook vitality via \textit{exponential moving average (EMA)} and \textit{dead code expiration}. Together, these designs enable the codebook to evolve with sufficient coverage and separability, as further evidenced by the codebook geometry analysis in Section~\ref{pdd}.

\mypara{Swin Decoder.}
The VQ-SwinTTS decoder mirrors the encoder, replacing patch embedding with scale-invariant Swin blocks to reconstruct representations across all hierarchical scales. We first decompose the multi-scale quantized outputs $\mathbf{Z}'_{\mathrm{ms}}$ into their constituent scales:
\begin{equation}
\{\mathbf{Z}^{'}_{(1)}, \mathbf{Z}^{'}_{(2)},..., \mathbf{Z}^{'}_{(H)}\} = \mathbf{Z}^{'}_{\mathrm{ms}}, \mathbf{Z}^{'}_{(h)} \in \mathbb{R}^{N_{h} \times D},
\end{equation}
where the patch counts $N_{1} = N_{2} = 2N_3 = \dots = 2^{H-1}N_{H}$.

The decoder takes the coarsest representation $\mathbf{Z}'_{(H)}$ as the initial input, and temporal resolution is incrementally restored through hierarchical upsampling. To ensure independent codebook optimization, there are no skip connections between the encoder and decoder. At each scale $h$, the decoder performs cross-attention, employing the codebook-quantized tokens $\mathbf{Z}'_{(h)}$ as keys and values, while the decoder's hidden features serve as Queries. Finally, the processed representations are projected back to the temporal dimension to yield the reconstructed time series.

\mypara{Training Objective.}
We optimize the model by minimizing a composite objective: a masked mean absolute error (MAE) to handle padded and missing values, alongside a \textit{commitment loss} that anchors encoder outputs to their target codebook vectors:
\begin{equation}
\mathcal{L}_{total} = \underbrace{\frac{1}{|\mathcal{M}|} \sum_{i \in \mathcal{M}} |x_i - \hat{x}_i|}_{\mathcal{L}_{rec}} + \beta \underbrace{\| \mathbf{W}_{in}\mathbf{z}_e - \text{sg}[q_k] \|^2_2}_{\mathcal{L}_{commit}},
\end{equation}
where $\mathcal{M}$ denotes the set of valid indices, $\hat{x}$ is the reconstructed series, $\text{sg}[\cdot]$ is the stop-gradient operator, and $\beta$ is the commitment weight. This strategy enables VQ-SwinTTS to learn a compact yet expressive dictionary of cross-scale temporal patterns. The resulting shared codebook space serves as a crucial foundation for defining series-level pattern equilibrium.

\begin{algorithm}[t]
\small
\renewcommand{\baselinestretch}{0.75}\selectfont
\caption{Pattern-Uniform Corpus Construction}
\label{alg:pattern_space}
\begin{algorithmic}[1]
\Require Learned Codebook $\mathbf{C} \in \mathbb{R}^{K \times d}$, Target Bins $M$.
\Ensure Pattern Space Parameters $\Theta = (\mu, \mathbf{V}, \mathbf{b}, \mathcal{E}, \mathcal{T})$.

\State \textbf{Stage 1: Spectral Rotation (KLT)}
\State Calculate centroid $\mu \leftarrow \text{mean}(\mathbf{C})$ and covariance $\Sigma \leftarrow \text{Cov}(\mathbf{C})$. \Comment{Eq. \ref{eq:covariance}}
\State Solve $\Sigma \mathbf{V} = \mathbf{V}\Lambda$ to get eigenvectors $\mathbf{V}$ sorted by $\lambda_i$.
\State Project codebook: $\tilde{\mathbf{C}} \leftarrow (\mathbf{C} - \mu)\mathbf{V}$.

\State \textbf{Stage 2: Optimal Resolution Allocation}
\State Init $\mathbf{b} \leftarrow \mathbf{1} \in \mathbb{Z}^d$ and priorities $\mathbf{w} \leftarrow [\sqrt{\lambda_1}, \dots, \sqrt{\lambda_d}]$.
\While{$\prod b_i < M$}
    \State $j^* \leftarrow \arg\max_j (\mathbf{w}_j / \mathbf{b}_j)$; \quad $b_{j^*} \leftarrow b_{j^*} + 1$ \Comment{Eq. \ref{eq:greedy_allocation}}
\EndWhile

\State \textbf{Stage 3: Density-Balanced Grid Construction}
\For{$i = 1$ to $d$}
    \State $\mathcal{E}_i \leftarrow \text{Quantile}(\tilde{\mathbf{C}}_{:, i}, \text{linspace}(0, 1, b_i+1))$ \Comment{Eq. \ref{eq:max_entropy}}
\EndFor

\State \textbf{Stage 4: Bin Unification}
\State Get keys $\mathbf{h} \leftarrow \text{MixedRadix}(\text{Digitize}(\tilde{\mathbf{C}}, \mathcal{E}), \mathbf{b})$ and histogram $H$.
\State Init map $\mathcal{T}$, $acc \leftarrow 0$, $u \leftarrow 0$. \Comment{Implements mapping $\psi$}
\For{$j = 0$ to $\prod b_i - 1$} \Comment{Merge adjacent grids}
    \State $\mathcal{T}[j] \leftarrow u$; \quad $acc \leftarrow acc + H[j]$
    \If{$acc \ge \frac{K}{M}(u + 1)$ \textbf{and} $u < M - 1$} $u \leftarrow u + 1$ \EndIf
\EndFor

\State \Return $\Theta = (\mu, \mathbf{V}, \mathbf{b}, \mathcal{E}, \mathcal{T})$
\end{algorithmic}
\end{algorithm}

\subsection{Uniform Pattern Binning}

While VQ-SwinTTS extracts patch-level multi-scale features and forms series-level semantics through codebook-based aggregation, these representations remain non-uniform and are not directly suitable for corpus-level balancing over large-scale real-world data. Raw series occupy learned codes unevenly, and aggregated dimensions have unequal variance. To address this issue, we further construct a scalable pattern space and perform \textit{uniform pattern binning} over the full raw corpus, resulting in a set of $M$ discrete \textit{pattern bins} $\mathcal{U}=\{u_1,\dots,u_M\}$ for subsequent learnability synchronization. The key idea is to transform non-uniform series-level semantics into comparable and finite semantic units that support large-scale partitioning and reweighting. Concretely, each series is assigned a bin ID by (i) encoding it into codebook vectors, (ii) aggregating them into a series-level embedding, and (iii) discretizing the embedding into a finite state space, as summarized in Algorithm~\ref{alg:pattern_space}.

\mypara{Spectral Rotation and Variance Alignment.}
To decorrelate codebook features and redistribute their spectral energy, we apply \textit{spectral rotation} via the Karhunen-Loève Transform~(KLT)~\cite{gastpar2006distributed}, which is empirically equivalent to principal component analysis~(PCA). Given the codebook centroid $\mu = \frac{1}{K}\sum q_k$, we compute the empirical covariance matrix $\Sigma$:
\begin{equation}
\Sigma = \frac{1}{K-1} \sum_{k=1}^K (q_k - \mu)(q_k - \mu)^\top, \label{eq:covariance}
\end{equation}
where $q_k \in \mathbb{R}^d$. Solving the eigenvalue problem $\Sigma \mathbf{V} = \mathbf{V} \Lambda$ yields eigenvectors sorted by spectral energy $\lambda_1 \ge \dots \ge \lambda_d$. The codebook is then projected into this variance-ordered basis as $\tilde{\mathbf{C}} = (\mathbf{C} - \mu)\mathbf{V}$, so that higher-variance components are prioritized in subsequent discretization. Figure~\ref{fig:codebook_analysis} reports the resulting spectral and geometric structure.

\mypara{Optimal Resolution Allocation}
A core challenge is determining how finely to divide each rotated dimension $i$. Rather than relying on simple heuristics, we formulate the allocation of bin granularity $b_i$ grounded in \textit{high-resolution quantization theory} \cite{gersho2012vector, huang1963block}. 

For a principal component with variance $\lambda_i$, its dynamic range is proportional to its standard deviation $\sigma_i = \sqrt{\lambda_i}$. If we allocate $b_i$ bins to this dimension, the spatial quantization step size (or bin width) is $\Delta_i \propto \sqrt{\lambda_i} / b_i$. To ensure uniform geometric fidelity across the latent space, our goal is to minimize the maximum spatial quantization error (\ie min-max $L_\infty$ distortion) under a strict total capacity constraint $M$.

Mathematically, this min-max optimization translates to equalizing the marginal step sizes across all dimensions. By integrating the bin width $\Delta_i \propto \sqrt{\lambda_i} / b_i$ with respect to the bin count $b_i$, we derive the resource allocation objective $J(\mathcal{B})$:
\begin{equation} \label{eq:allocation_objective}
\max J(\mathcal{B}) = \sum_{i=1}^{d} \sqrt{\lambda_i} \ln b_i, \quad \text{s.t.} \prod_{i=1}^{d} b_i \le M.
\end{equation}
We solve this via a greedy allocation strategy (Algorithm~\ref{alg:pattern_space}), iteratively incrementing the bin count $b_j$ that yields the maximum discrete marginal gain:
\begin{equation} \label{eq:greedy_allocation}
j^{*} = \arg\max_j \big(\sqrt{\lambda_j} \ln(b_j+1) - \sqrt{\lambda_j} \ln b_j \big) \approx \arg\max_j \frac{\sqrt{\lambda_j}}{b_j}.
\end{equation}
This strategy elegantly guarantees that the dimension with the currently largest quantization error ($\sqrt{\lambda_j}/b_j$) is selected for further refinement. It effectively performs soft dimensionality reduction by collapsing noise-dominated dimensions ($b_i = 1$) while allocating high resolution to principal components, preserving essential morphological diversity \cite{gersho1979asymptotically}.

\mypara{Density-Balanced Grid and Bin Unification.}
To ensure robustness against outliers, partition boundaries $\mathcal{E}_i$ are defined using the \textit{empirical cumulative distribution function~(ECDF)} of the projected values $\tilde{\mathbf{C}}_{:,i}$. Boundaries are selected at quantiles $j/b_i$ for $j=1,\dots,b_i-1$, giving equal codebook occupancy per interval:
\begin{equation}
P(e_{i,j-1} < \tilde{q}^{(i)} \le e_{i,j}) = \frac{1}{b_i}. \label{eq:max_entropy}
\end{equation}
The Cartesian product of these partitions forms a grid of size $S_{grid} = \prod b_i$. To align with the exact capacity $M$, we apply \textit{bin unification}. Grid indices are flattened via Mixed Radix Encoding and merged through a surjective mapping $\psi$ based on occupancy density, ensuring an approximately uniform probability mass $\sum_{\psi(h(g))=u} P(g) \approx 1/M$ for each final bin $u \in \mathcal{U}$.

\mypara{Connectivity to the Training Pipeline.} 
For any raw input series, its multi-scale representations from VQ-SwinTTS are deterministically projected onto $\mathbf{V}$ and assigned to a semantic bin $u_m \in \mathcal{U}$. While applying standard Inverse Frequency Weighting on these bins achieves \textit{distribution equilibrium} by flattening the long-tail distribution, it fatally ignores the intrinsic \textit{learnability discrepancy}: simple patterns converge vastly faster than complex ones. Thus, Stage~2 yields a pattern-balanced candidate corpus, but still treats all bins as equally learnable. This remaining optimization gap is resolved in the final stage via GRPO-based synchronization.

\subsection{Learnability Synchronization via GRPO}
Stage~3 starts from the pattern-balanced bins $\mathcal{U}$ and learns weights for bins with different convergence speeds $r_{i,t}$.
As illustrated in Figure~\ref{fig:main}, we propose \textit{learnability synchronization}, an offline bi-level optimization framework that explicitly profiles bin-wise learnability and uses it to guide reweighting. To ensure scalability, the framework is decoupled into two sequential phases: offline profiling and policy optimization, detailed below and in Eqs.~\eqref{eq:advantage}--\eqref{eq:grpo_loss}.

\mypara{Phase 1: Offline Learnability Profiling (Inner Loop).}
Pre-training universal forecasting models incurs prohibitive computational overhead; training a lightweight proxy like ChronosBolt-Small takes $\sim$6 GPU-hours, while TimeMoE-Base requires 3 days. Thus, extracting online dynamic feedback to synchronously update the GRPO policy during training is computationally infeasible. To circumvent this bottleneck, we conduct an offline profiling session to efficiently estimate bin-wise convergence speeds $r_{i,t}$.
\begin{itemize}[leftmargin=*, nosep]
\item \textit{Uniform probing:} We initialize the forecasting model with {uniform sampling weights} $\mathbf{w}_{uni}$ and train it for a complete session.
\item \textit{Convergence speed extraction:} Rather than relying on a single static metric, we record the validation loss trajectory across multiple training stages $t$. The intrinsic convergence speed $r_{i,t}$ is computed dynamically as the slope of this trajectory at specific checkpoints (\eg the reduction in MSE per epoch). To ensure robustness, we average $r_{i,t}$ across multiple random seeds.
\end{itemize}
This phase captures the non-stationary optimization dynamics, yielding a stage-aware "Learnability Profile" matrix $\mathbf{R} \in \mathbb{R}^{M \times T}$, where a higher $r_{i,t}$ indicates a pattern that converges faster at training stage $t$.

\mypara{Phase 2: GRPO Optimization (Outer Loop).}
With the pre-computed profile $\mathbf{R}$, we optimize the sampling weights to synchronize learning speeds. A naive closed-form inverse weighting ($w_i \propto 1/r_i$) is unstable in large-scale pre-training: the speeds are stage-dependent and estimated from stochastic, parameter-coupled mini-batches. To address this stochastic local context, we treat the weights as a policy $\pi_\theta$ and employ \textit{group relative policy optimization (GRPO)}, which uses group-normalized rewards.

\mypara{Policy Modeling}
For each bin $i$, we use a lightweight policy network $\pi_\theta$ that maps its embedding $v_i$ to a sampling logit. During optimization, instead of viewing all bins globally, we sample a subset $\mathcal{B}$. We introduce exploration noise to the sample weights:
\begin{equation}
\mu_i = f_\theta(v_i), \quad w_i = \text{Softmax}(\mu_i + \mathcal{N}(0, \sigma)), \quad \forall i \in \mathcal{B}.
\end{equation}

\mypara{Reward Design}
We define the \textit{Effective Learning Speed} of bin $i$ at stage $t$ as $p_{i,t} = w_i \cdot r_{i,t}$. The goal is to ensure that all bins within the current stochastic batch learn at a synchronized pace. Thus, the reward penalizes the deviation of a bin's speed from the local mean speed $\bar{p}_t$ within the sampled subset $\mathcal{B}$:
\begin{equation}
R_{i,t} = - \left( w_i r_{i,t} - \bar{p}_t \right)^2, \quad \text{where } \bar{p}_t = \frac{1}{|\mathcal{B}|} \sum_{j \in \mathcal{B}} w_j r_{j,t}.
\end{equation}
This reward mechanism automatically assigns lower weights to "fast-learner" bins and higher weights to "slow-learners" within highly variable local contexts, forcing dynamic synchronization.

\mypara{Group Update}
We employ GRPO to update $\theta$ efficiently. To reduce variance, we maintain a {group pool} $\mathcal{H}$ of historical weight-reward pairs. For a sampled group $\mathcal{G}$, we first compute the standardized advantage $A_i$ to normalize the reward scale:
\begin{equation} \label{eq:advantage}
A_i = \frac{R_i - \mathbb{E}_{j \in \mathcal{G}}[R_j]}{\text{Std}_{j \in \mathcal{G}}(R_j) + \epsilon}.
\end{equation}
Subsequently, we minimize the policy gradient objective with importance sampling clipping:
\begin{equation} \label{eq:grpo_loss}
\mathcal{L}_{\text{GRPO}} = -\mathbb{E}_{\mathcal{G} \sim \mathcal{H}} \bigg[ \frac{1}{|\mathcal{G}|} \sum_{i \in \mathcal{G}} \min \big( \rho_i A_i, \text{clip}(\rho_i, 1\!-\!\epsilon, 1\!+\!\epsilon) A_i \big) \bigg].
\end{equation}
Since $r_i$ is pre-computed, this outer loop does not involve expensive model training, allowing for thousands of rapid policy iterations to find the optimal weight $\mathbf{w}^*$. Consequently, comparable or better forecasting accuracy can be achieved with fewer pre-training tokens, as demonstrated in our experiments.

Overall, Stages 1--3 complete the QUALS pipeline, progressing from deep pattern extraction, to uniform pattern binning over the full raw corpus, and finally to learnability-aware reweighting. Together, they construct a corpus balanced in both pattern distribution and optimization dynamics.

\begin{table*}[t]
\setlength{\abovecaptionskip}{0.cm}
\setlength{\belowcaptionskip}{0.cm}
\centering
\footnotesize                    
\setlength{\tabcolsep}{3pt}      
\renewcommand{\arraystretch}{0.85}
\caption{Detailed setting configurations for the QUALS framework.}
\label{tab:hyperparameters}
\resizebox{0.96\linewidth}{!}{
\begin{tabular*}{\linewidth}{@{\extracolsep{\fill}} ll ll ll @{}}
\toprule
\multicolumn{2}{c}{\textbf{Stage 1: VQ-SwinTTS Backbone}} & \multicolumn{2}{c}{\textbf{Stage 2: Pattern Binning \& Sampling}} & \multicolumn{2}{c}{\textbf{Stage 3: GRPO Synchronization}} \\
\cmidrule{1-2} \cmidrule{3-4} \cmidrule{5-6}
\textbf{Setting} & \textbf{Value} & \textbf{Setting} & \textbf{Value} & \textbf{Setting} & \textbf{Value} \\
\midrule
Input Sequence Length ($L$) & 4096 & Valid Pattern Bins & 2,499 & Embedding Dimension($d$) & 32 \\
Patch Size ($P$) & 32 & Out-of-Bound Bins (Empty) & 1 & Hidden Dimension & 64 \\
Stride Size ($S$) & 16 & Sampled Sequence Length & 4096 & Learning Rate & 1e-5 \\
Codebook Size ($K$) & 2048 & Train Series per Bin & 8,000 & Subset Size per Group ($|\mathcal{B}|$) & 480 \\
Codebook Dimension ($d$) & 32 & Pre-training Corpus Volume & 81.89B & Number of Groups ($G$) & 128 \\
Hidden Dimension ($D$) & 256 & Val Series per Bin & 160 & Clip Epsilon ($\epsilon$) & 0.2 \\
Attention Heads & $[2, 4, 8, 16]$ & Validation Corpus Volume & 1.64B & Exploration Ratio & 0.2 \\
\bottomrule
\end{tabular*}}
\end{table*}

\subsection{Implementation Detail}
\label{pipeline}
To synthesize QUALS, we outline the implementation pipeline, and the relevant settings are shown in Table~\ref{tab:hyperparameters}:
\begin{itemize}[leftmargin=*]
\item \textbf{Stage 1:} The VQ-SwinTTS backbone is trained on the BLAST corpus comprising 81.1B tokens ($\sim$20 GPU-hours), solely for representation learning so as to reduce model bias. No subsequent pattern binning or corpus balancing is performed on BLAST.

\item \textbf{Stage 2:} 
We first apply the frozen VQ-SwinTTS to the massive 239B raw GIFT-Eval Pretrain corpus ($\sim$2 GPU-hours). Using PCA and ECDF, the codebook space is partitioned into 2,499 valid pattern bins (the 2,500th remains empty due to empirical boundary conditions). For each sequence, we extract its multi-scale representations, linearly aggregate and project them into the global space, and assign a bin index.
Then, 
based on the global bin assignments, we construct the candidate corpus by sampling sequences of fixed length $L=4096$. Specifically, 8,000 series per bin are selected for pre-training (yielding 81.89B points) and 160 series per bin for validation (1.64B points). The data sampling and disk I/O operations are executed efficiently ($\sim$10 hours).
The 8,000-per-bin quota is used to construct the Stage 2 candidate corpus, while Stage 3 determines the final sampling probabilities.
\item \textbf{Stage 3:} A lightweight proxy, ChronosBolt-\allowbreak Small, is trained uniformly on the candidate corpus to extract convergence speeds ($\sim$6 GPU-hours). Offline GRPO weight optimization then converges rapidly ($\sim$1 GPU-hour).
Stage~3 outputs normalized bin weights $\mathbf{w}^*\in\mathbb{R}^{M}$ over the 2,499 valid bins.
\end{itemize}

\begin{table*}[t]
\setlength{\abovecaptionskip}{0.cm}
\setlength{\belowcaptionskip}{0.cm}
\caption{
Performance of models retrained from scratch on \textbf{BLAST (B)} or \textbf{QUALS (Q)} versus their pretrained counterparts.
Lower \textbf{MAE/MSE} is better.
$s/b/l$ denote small/base/large.
Models with superior or equal performance are highlighted in {\color{MyRed2}\textbf{red}}.
}

\setlength\tabcolsep{1.pt}
\renewcommand{\arraystretch}{0.85}
\resizebox{1\linewidth}{!}{
\begin{tabular}{rr|cccccc|cccccc|cccccc|cccccc|cccccc|cccccc}
\toprule[1.5pt]
\multicolumn{2}{l}{\textbf{Models}} & \multicolumn{2}{|l}{\textbf{TimeMoE}$_{l}^{\textbf{Q}}$}& \multicolumn{2}{l}{\textbf{TimeMoE}$_{l}^\textbf{B}$} & \multicolumn{2}{l|}{\textbf{TimeMoE}$_{l}$} & \multicolumn{2}{l}{\textbf{TimeMoE}$_{b}^{\textbf{Q}}$}& \multicolumn{2}{l}{\textbf{TimeMoE}$_{b}^\textbf{B}$} & \multicolumn{2}{l|}{\textbf{TimeMoE}$_{b}$} & \multicolumn{2}{l}{\textbf{MOIRAI}$_{l}^{\textbf{Q}}$}& \multicolumn{2}{l}{\textbf{MOIRAI}$_{l}^\textbf{B}$} & \multicolumn{2}{l|}{\textbf{MOIRAI}$_{l}$} & \multicolumn{2}{l}{\textbf{MOIRAI}$_{b}^{\textbf{Q}}$}& \multicolumn{2}{l}{\textbf{MOIRAI}$_{b}^\textbf{B}$} & \multicolumn{2}{l|}{\textbf{MOIRAI}$_{b}$} & \multicolumn{2}{l}{\textbf{Chronos}$_{b}^{\textbf{Q}}$}& \multicolumn{2}{l}{\textbf{Chronos}$_{b}^\textbf{B}$} & \multicolumn{2}{l|}{\textbf{Chronos}$_{b}$} & \multicolumn{2}{l}{\textbf{Chronos}$_{s}^{\textbf{Q}}$} & \multicolumn{2}{l}{\textbf{Chronos}$_{s}^{\textbf{B}}$} & \multicolumn{2}{l}{\textbf{Chronos}$_{s}$} \\
\cmidrule(r){3-4} \cmidrule(r){5-6} \cmidrule(r){7-8} \cmidrule(r){9-10} \cmidrule(r){11-12} \cmidrule(r){13-14} \cmidrule(r){15-16} \cmidrule(r){17-18} \cmidrule(r){19-20} \cmidrule(r){21-22} \cmidrule(r){23-24} \cmidrule(r){25-26}\cmidrule(r){27-28} \cmidrule(r){29-30} \cmidrule(r){31-32} \cmidrule(r){33-34}
\cmidrule(r){35-36} \cmidrule(r){37-38}
\multicolumn{2}{l|}{{Metrics}} & MSE & MAE & MSE & MAE & MSE & MAE & MSE & MAE & MSE & MAE & MSE & MAE & MSE & MAE & MSE & MAE & MSE & MAE & MSE & MAE & MSE & MAE & MSE & MAE & MSE & MAE & MSE & MAE & MSE & MAE & MSE & MAE & MSE & MAE & MSE & MAE\\
\midrule
\multirow{5}{*}{\rotatebox{90}{ETTh1}} & 96
&{\color{MyRed2}\textbf{.340}}&{\color{MyRed2}\textbf{.373}}& .348 & .375 & .350 & .382 & {\color{MyRed2}\textbf{.342}}&{\color{MyRed2}\textbf{.373}}& .352 & .376 & .357 & .381
&.364&{\color{MyRed2}\textbf{.379}}& {\color{MyRed2}\textbf{.359}} & .383 & .381 & .388
&.369&{\color{MyRed2}\textbf{.380}}&{\color{MyRed2}\textbf{.362}} & .384 & .376 & .392
&{\color{MyRed2}\textbf{.347}}&{\color{MyRed2}\textbf{.370}}& .357 & .375 & .384 & .379
&{\color{MyRed2}\textbf{.342}}&{\color{MyRed2}\textbf{.371}}& .359 & .376 & .394 &  .381 \\
& 192
&{\color{MyRed2}\textbf{.376}}&{\color{MyRed2}\textbf{.399}}& .381 & {\color{MyRed2}\textbf{.399}} & .388 & .412
&{\color{MyRed2}\textbf{.375}}&{\color{MyRed2}\textbf{.397}}& .389 & .401 & .384 & .404
&.404&{\color{MyRed2}\textbf{.401}}& {\color{MyRed2}\textbf{.395}} & .404 & .434 & .415
&.406&.406& {\color{MyRed2}\textbf{.400}} & {\color{MyRed2}\textbf{.405}} & .412 & .413
&{\color{MyRed2}\textbf{.390}}&{\color{MyRed2}\textbf{.396}}& .397 & .401 & .441 & .412
&{\color{MyRed2}\textbf{.383}}&{\color{MyRed2}\textbf{.396}}& .403 & .402 & .455 & .414 \\
& 336
&{\color{MyRed2}\textbf{.402}}&{\color{MyRed2}\textbf{.416}}& .409 & .424 & .411 & .430
&{\color{MyRed2}\textbf{.395}}&{\color{MyRed2}\textbf{.410}}& .408 & .419 & .411 & .434
&.424&{\color{MyRed2}\textbf{.411}}& {\color{MyRed2}\textbf{.411}} & .416 & .495 & .445
&.426&.416& {\color{MyRed2}\textbf{.416}} & .420 & .433 & .428
&{\color{MyRed2}\textbf{.412}}&{\color{MyRed2}\textbf{.409}}& .422 & .417 & .475 & .430
&{\color{MyRed2}\textbf{.406}}&{\color{MyRed2}\textbf{.409}}& .431 & .416 & .499 & .444 \\
& 720
&.437&{\color{MyRed2}\textbf{.447}}& .447 & .451 & {\color{MyRed2}\textbf{.427}} & .455
&{\color{MyRed2}\textbf{.421}}&{\color{MyRed2}\textbf{.441}}& .450 & .455 & .449 & .477
&{\color{MyRed2}\textbf{.413}}&{\color{MyRed2}\textbf{.427}}& .420 & .430 & .611 & .510
&{\color{MyRed2}\textbf{.428}}&{\color{MyRed2}\textbf{.429}}& .430 & .439 & .447 & .444
&{\color{MyRed2}\textbf{.415}}&{\color{MyRed2}\textbf{.427}}& .460 & .443 & .472 & .446
&{\color{MyRed2}\textbf{.398}}&{\color{MyRed2}\textbf{.421}}& .449 & .439 & .520 & .476 \\
\rowcolor{tabhighlight}\cellcolor{white}
& AVG
&{\color{MyRed2}\textbf{.389}}&{\color{MyRed2}\textbf{.408}}& .396 & .412 & .394 & .419
&{\color{MyRed2}\textbf{.383}}&{\color{MyRed2}\textbf{.405}}& .399 & .412 & .400 & .424
&.401&{\color{MyRed2}\textbf{.404}}& {\color{MyRed2}\textbf{.396}} & .408 & .480 & .439
&.407&{\color{MyRed2}\textbf{.408}}& {\color{MyRed2}\textbf{.402}} & .412 & .417 & .419
&{\color{MyRed2}\textbf{.391}}&{\color{MyRed2}\textbf{.401}}& .409 & .409 & .443 & .416
&{\color{MyRed2}\textbf{.382}}&{\color{MyRed2}\textbf{.399}}& .410 & .408 & .467 & .428 \\

\midrule

\multirow{5}{*}{\rotatebox{90}{ETTh2}} & 96
&{\color{MyRed2}\textbf{.273}}&.330& .276 & {\color{MyRed2}\textbf{.329}} & .302 & .354
&{\color{MyRed2}\textbf{.265}}&{\color{MyRed2}\textbf{.328}}& .285 & .332& .305 & .359
&{\color{MyRed2}\textbf{.278}}&{\color{MyRed2}\textbf{.320}}& .288 & .325 & .296 & .330
&{\color{MyRed2}\textbf{.280}}&{\color{MyRed2}\textbf{.319}}& .284 & .324 & .294 & .325
&{\color{MyRed2}\textbf{.275}}&{\color{MyRed2}\textbf{.319}}& .282 & .321 & .289 & .330
&{\color{MyRed2}\textbf{.275}}&{\color{MyRed2}\textbf{.319}}& .281 & .326 & .282 & .328 \\
& 192
&{\color{MyRed2}\textbf{.342}}&.380& .345& {\color{MyRed2}\textbf{.376}} & .364 & .385
&{\color{MyRed2}\textbf{.331}}&{\color{MyRed2}\textbf{.378}}& .348 & {\color{MyRed2}\textbf{.378}} & .351 & .386
&{\color{MyRed2}\textbf{.338}}&{\color{MyRed2}\textbf{.362}}& .353 & .370 & .361 & .371
&{\color{MyRed2}\textbf{.344}}&{\color{MyRed2}\textbf{.362}}& .348 & .369 & .365 & .375
&{\color{MyRed2}\textbf{.342}}&{\color{MyRed2}\textbf{.364}}& .356 & .369 & .359 & .369
&{\color{MyRed2}\textbf{.341}}&{\color{MyRed2}\textbf{.365}}& .353 & .371 & .354 & .373 \\
& 336
&{\color{MyRed2}\textbf{.366}}&{\color{MyRed2}\textbf{.401}}& .384 & .416 & .417 & .425
&{\color{MyRed2}\textbf{.360}}&{\color{MyRed2}\textbf{.398}}& .372 & .405 & .391 & .418
&{\color{MyRed2}\textbf{.364}}&.387& .369 & {\color{MyRed2}\textbf{.382}} & .390 & .390
&.372&.393& {\color{MyRed2}\textbf{.367}} & {\color{MyRed2}\textbf{.386}} & .376 & .390
&{\color{MyRed2}\textbf{.378}}&{\color{MyRed2}\textbf{.393}}& {\color{MyRed2}\textbf{.378}} & .397 & .399 & .400
&{\color{MyRed2}\textbf{.372}}&{\color{MyRed2}\textbf{.392}}& .387 & .403 & .416 & .410 \\
& 720
&{\color{MyRed2}\textbf{.401}}&{\color{MyRed2}\textbf{.438}}& .442 & .470 & .537 & .496
&{\color{MyRed2}\textbf{.395}}&{\color{MyRed2}\textbf{.433}}& .419 & .452 & .419 & .454
&{\color{MyRed2}\textbf{.382}}&.412& .387 & {\color{MyRed2}\textbf{.406}} & .423 & .418
&{\color{MyRed2}\textbf{.382}}&{\color{MyRed2}\textbf{.404}}& .387 & .410 & .416 & .433
&{\color{MyRed2}\textbf{.399}}&{\color{MyRed2}\textbf{.419}}& .403 & .424 & .420 & .425
&{\color{MyRed2}\textbf{.394}}&{\color{MyRed2}\textbf{.417}}& .411 & .430 & .428 & .431 \\
\rowcolor{tabhighlight}\cellcolor{white}
& AVG
&{\color{MyRed2}\textbf{.345}}&{\color{MyRed2}\textbf{.387}}& .361 & .397& .405 & .415
&{\color{MyRed2}\textbf{.338}}&{\color{MyRed2}\textbf{.384}}& .356& .391 & .366 & .404
&{\color{MyRed2}\textbf{.341}}&{\color{MyRed2}\textbf{.370}}& .349 & {\color{MyRed2}\textbf{.370}} & .367 & .377
&{\color{MyRed2}\textbf{.344}}&{\color{MyRed2}\textbf{.369}}& .346 & .372 & .362 & .382
&{\color{MyRed2}\textbf{.349}}&{\color{MyRed2}\textbf{.374}}& .355 & .377 & .366 & .381
&{\color{MyRed2}\textbf{.346}}&{\color{MyRed2}\textbf{.373}}& .358 & .382 & .370 & .385 \\

\midrule
\multirow{5}{*}{\rotatebox{90}{ETTm1}} & 96
&{\color{MyRed2}\textbf{.288}}&{\color{MyRed2}\textbf{.332}}& .327 & .343& .309 & .357
&{\color{MyRed2}\textbf{.300}}&{\color{MyRed2}\textbf{.340}}& .334 & .350 & .338 & .368
&{\color{MyRed2}\textbf{.341}}&{\color{MyRed2}\textbf{.347}}& .355 & .355 & .380 & .361
&{\color{MyRed2}\textbf{.334}}&{\color{MyRed2}\textbf{.349}}& .348 & .354 & .363 & .356
&.321&.334& {\color{MyRed2}\textbf{.310}} & {\color{MyRed2}\textbf{.327}} & .331 & .333
&{\color{MyRed2}\textbf{.309}}&.335& .314 & {\color{MyRed2}\textbf{.331}} & .328 & .332 \\
& 192
&{\color{MyRed2}\textbf{.328}}&{\color{MyRed2}\textbf{.362}}& .368 & .378 & .346 & .381
&{\color{MyRed2}\textbf{.337}}&{\color{MyRed2}\textbf{.366}}& .388 & .386 & .353 & .388
&{\color{MyRed2}\textbf{.378}}&{\color{MyRed2}\textbf{.370}}& .388 & .380 & .412 & .383
&.{\color{MyRed2}\textbf{366}}&{\color{MyRed2}\textbf{.369}}& .385 & .378 & .388 & .375
&.377&.373& {\color{MyRed2}\textbf{.363}} & {\color{MyRed2}\textbf{.360}} & .386 & .365
&{\color{MyRed2}\textbf{.355}}&.366& .364 & {\color{MyRed2}\textbf{.365}} & .365 & .384 \\
& 336
&{\color{MyRed2}\textbf{.356}}&{\color{MyRed2}\textbf{.385}}& .373 & .396 & .373 & .408
&{\color{MyRed2}\textbf{.364}}&{\color{MyRed2}\textbf{.387}}& .400 & .412 & .381 & .413
&{\color{MyRed2}\textbf{.398}}&{\color{MyRed2}\textbf{.385}}& .399 & .387 & .436 & .400
&{\color{MyRed2}\textbf{.387}}&{\color{MyRed2}\textbf{.384}}& .410 & .394 & .416 & .392
&{\color{MyRed2}\textbf{.405}}&.387& .410 & .387 & .408 & {\color{MyRed2}\textbf{.382}}
&{\color{MyRed2}\textbf{.382}}&{\color{MyRed2}\textbf{.390}}& .417 & .391 & .425 & .391 \\
& 720
&{\color{MyRed2}\textbf{.410}}&{\color{MyRed2}\textbf{.421}}& .445 & .438 & .475 & .477
&{\color{MyRed2}\textbf{.415}}&{\color{MyRed2}\textbf{.419}}& .457 & .451 & .504 & .493
&{\color{MyRed2}\textbf{.428}}&{\color{MyRed2}\textbf{.407}}& .429 & .413 & .462 & .420
&{\color{MyRed2}\textbf{.423}}&{\color{MyRed2}\textbf{.409}}& .448 & .416 & .460 & .418
&{\color{MyRed2}\textbf{.458}}&{\color{MyRed2}\textbf{.422}}& .477 & .427 & .503 & .430
&{\color{MyRed2}\textbf{.435}}&{\color{MyRed2}\textbf{.433}} & .521  &.452& .525 &.445 \\
\rowcolor{tabhighlight}\cellcolor{white}
& AVG
&{\color{MyRed2}\textbf{.346}}&{\color{MyRed2}\textbf{.375}}& .378 & .388 & .375 & .405
&{\color{MyRed2}\textbf{.354}}&{\color{MyRed2}\textbf{.378}}& .394 & .399 & .394 & .415
&{\color{MyRed2}\textbf{.386}}&{\color{MyRed2}\textbf{.377}}& .392 & .383 & .422 & .391
&{\color{MyRed2}\textbf{.377}}&{\color{MyRed2}\textbf{.378}}& .397 & .385 & .406 & .385
&{\color{MyRed2}\textbf{.390}}&.379& {\color{MyRed2}\textbf{.390}} & {\color{MyRed2}\textbf{.375}} & .407 & .377
&{\color{MyRed2}\textbf{.371}}&{\color{MyRed2}\textbf{.381}}& .404 & .385 & .411 & .388 \\

\midrule
\multirow{5}{*}{\rotatebox{90}{ETTm2}} & 96
&{\color{MyRed2}\textbf{.161}}&{\color{MyRed2}\textbf{.249}}& .180 & .259 & .197 & .286
&{\color{MyRed2}\textbf{.166}}&{\color{MyRed2}\textbf{.254}}& .181 & .260 & .201 & .291
&{\color{MyRed2}\textbf{.174}}&{\color{MyRed2}\textbf{.246}}& .192 & .259 & .211 & .274
&{\color{MyRed2}\textbf{.186}}&{\color{MyRed2}\textbf{.258}}& .194 & .265 & .205 & .273
&{\color{MyRed2}\textbf{.173}}&.245& .175 & .249 & .177 & {\color{MyRed2}\textbf{.244}}
&.182&.251& {\color{MyRed2}\textbf{.180}} & {\color{MyRed2}\textbf{.248}} & {\color{MyRed2}\textbf{.180}} & .251 \\
& 192
&{\color{MyRed2}\textbf{.212}}&{\color{MyRed2}\textbf{.288}}& .245 & .305 & .250 & .322
&{\color{MyRed2}\textbf{.218}}&{\color{MyRed2}\textbf{.294}}& .247 & .307 & .258 & .334
&{\color{MyRed2}\textbf{.230}}&{\color{MyRed2}\textbf{.284}}& .256 & .302 & .281 & .318
&{\color{MyRed2}\textbf{.242}}&{\color{MyRed2}\textbf{.297}}& .257 & .304 & .275 & .316
&{\color{MyRed2}\textbf{.235}}&{\color{MyRed2}\textbf{.289}}& .242 & .290 & .251 & .293
&.251&.297& {\color{MyRed2}\textbf{.243}} & {\color{MyRed2}\textbf{.292}} & .251 & .298 \\
& 336
&{\color{MyRed2}\textbf{.260}}&{\color{MyRed2}\textbf{.324}}& .283 & .338 & .337 & .375
&{\color{MyRed2}\textbf{.268}}&{\color{MyRed2}\textbf{.329}}& .293 & .344 & .324 & .373
&{\color{MyRed2}\textbf{.279}}&{\color{MyRed2}\textbf{.318}}& .289 & .329 & .341 & .355
&{\color{MyRed2}\textbf{.294}}&{\color{MyRed2}\textbf{.331}}& .301 & .342 & .329 & .350
&{\color{MyRed2}\textbf{.293}}&.327& .299 & {\color{MyRed2}\textbf{.326}} & .305 & .327
&.311&.338& {\color{MyRed2}\textbf{.302}} & {\color{MyRed2}\textbf{.331}} & .315 & .338 \\
& 720
&{\color{MyRed2}\textbf{.338}}&{\color{MyRed2}\textbf{.378}}& .364& .392 & .480 & .461
&{\color{MyRed2}\textbf{.350}}&{\color{MyRed2}\textbf{.385}}& .376& .396 & .488 & .464
&{\color{MyRed2}\textbf{.352}}&{\color{MyRed2}\textbf{.366}}& .372 & .384 & .428 & .428
&{\color{MyRed2}\textbf{.373}}&{\color{MyRed2}\textbf{.381}}& .387 & .396 & .437 & .411
&.397&.394& {\color{MyRed2}\textbf{.394}} & {\color{MyRed2}\textbf{.387}} & .419 & .394
&{\color{MyRed2}\textbf{.405}}&{\color{MyRed2}\textbf{.396}}& .406 & {\color{MyRed2}\textbf{.396}} & .421 & .403 \\
\rowcolor{tabhighlight}\cellcolor{white}
& AVG
&{\color{MyRed2}\textbf{.243}}&{\color{MyRed2}\textbf{.310}}& .268 & .323& .316 & .361
&{\color{MyRed2}\textbf{.251}}&{\color{MyRed2}\textbf{.316}}& .274 & .326 & .317 & .365
&{\color{MyRed2}\textbf{.259}}&{\color{MyRed2}\textbf{.303}}& .277 & .318 & .315 & .343
&{\color{MyRed2}\textbf{.274}}&{\color{MyRed2}\textbf{.317}}& .284 & .326 & .311 & .337
&{\color{MyRed2}\textbf{.275}}&{\color{MyRed2}\textbf{.313}}& .277 & {\color{MyRed2}\textbf{.313}} & .288 & .314
&.287&.321& {\color{MyRed2}\textbf{.282}} & {\color{MyRed2}\textbf{.316}} & .291 & .330 \\

\midrule

\multirow{5}{*}{\rotatebox{90}{Weather}} & 96
&{\color{MyRed2}\textbf{.156}}&{\color{MyRed2}\textbf{.203}}& .161 & .209 & .159 & .213
&{\color{MyRed2}\textbf{.156}}&{\color{MyRed2}\textbf{.205}}& .163 & .213 & .160 & .214
&{\color{MyRed2}\textbf{.167}}&.201& .168 & {\color{MyRed2}\textbf{.200}} & .278 & .376
&{\color{MyRed2}\textbf{.170}}&.203& .171 & {\color{MyRed2}\textbf{.202}} & .220 & .217
&{\color{MyRed2}\textbf{.156}}&{\color{MyRed2}\textbf{.191}}& .163 & .197 & .177 & .210
&{\color{MyRed2}\textbf{.157}}&{\color{MyRed2}\textbf{.193}}& .164 & .198 & .172 & .206 \\
& 192
&{\color{MyRed2}\textbf{.211}}&{\color{MyRed2}\textbf{.254}}& .217 & .261 & .215 & .266
&.211&{\color{MyRed2}\textbf{.257}}& .215 & .263 & {\color{MyRed2}\textbf{.210}} & .260
&{\color{MyRed2}\textbf{.212}}&{\color{MyRed2}\textbf{.244}}& .217 & .246 & .301 & .409
&.220&.251& {\color{MyRed2}\textbf{.218}} & {\color{MyRed2}\textbf{.247}} & .271 & .259
&{\color{MyRed2}\textbf{.202}}&{\color{MyRed2}\textbf{.236}}& .210 & .241 & .224 & .253
&{\color{MyRed2}\textbf{.205}}&{\color{MyRed2}\textbf{.240}}& .213 & .244 & .218 & .248 \\
& 336
&{\color{MyRed2}\textbf{.273}}&{\color{MyRed2}\textbf{.303}}& .276 & .304 & .291 & .322
&{\color{MyRed2}\textbf{.271}}&.303& .273 & {\color{MyRed2}\textbf{.297}} & .309 & .309
&{\color{MyRed2}\textbf{.264}}&{\color{MyRed2}\textbf{.283}}& .275 & .288 & .329 & .420
&{\color{MyRed2}\textbf{.278}}&{\color{MyRed2}\textbf{.289}}& {\color{MyRed2}\textbf{.278}} & .291 & .286 & .297
&{\color{MyRed2}\textbf{.259}}&.279& .264 & .282 & .260 & {\color{MyRed2}\textbf{.276}}
&{\color{MyRed2}\textbf{.261}}&{\color{MyRed2}\textbf{.282}}& .273 & .288 & .266 & {\color{MyRed2}\textbf{.282}} \\
& 720
&.354&.361& {\color{MyRed2}\textbf{.342}} & {\color{MyRed2}\textbf{.353}} & .415 & .400
&.353&.363& {\color{MyRed2}\textbf{.328}} & {\color{MyRed2}\textbf{.339}} & .418 & .405
&{\color{MyRed2}\textbf{.344}}&{\color{MyRed2}\textbf{.333}}& .375 & .351 & .370 & .463
&{\color{MyRed2}\textbf{.364}}&{\color{MyRed2}\textbf{.345}}& .370 & .350 & .373 & .354
&.341&{\color{MyRed2}\textbf{.330}}& {\color{MyRed2}\textbf{.339}} & .334 & .345 & .331
&{\color{MyRed2}\textbf{.348}}&{\color{MyRed2}\textbf{.339}}& .349 & .342 & .358 & .339 \\
\rowcolor{tabhighlight}\cellcolor{white}
& AVG
&{\color{MyRed2}\textbf{.248}}&{\color{MyRed2}\textbf{.280}}& .249 & .281 & .270 & .300
&.248&.282& {\color{MyRed2}\textbf{.244}} & {\color{MyRed2}\textbf{.278}} & .274 & .297
&{\color{MyRed2}\textbf{.247}}&{\color{MyRed2}\textbf{.265}}& .259 & .271 & .319 & .417
&{\color{MyRed2}\textbf{.258}}&{\color{MyRed2}\textbf{.272}}& .259 & {\color{MyRed2}\textbf{.272}} & .287 & .281
&{\color{MyRed2}\textbf{.240}}&{\color{MyRed2}\textbf{.259}}& .244 & .263 & .251 & .267
&{\color{MyRed2}\textbf{.243}}&{\color{MyRed2}\textbf{.263}}& .249 & .268 & .253 & .268 \\

\midrule
\rowc\rowcolor{blue!15}
\multicolumn{2}{c|}{\scalebox{1.1}{\textbf{\# Wins}}}
&\scalebox{1.1}{\color{MyRed2}\textbf{23}}&\scalebox{1.1}{\color{MyRed2}\textbf{22}}& 1& 4 & 1 & 0
& \scalebox{1.1}{\color{MyRed2}\textbf{22}}& \scalebox{1.1}{\color{MyRed2}\textbf{22}}& 2 & 4 & 1 & 0
&\scalebox{1.1}{\color{MyRed2}\textbf{21}}&\scalebox{1.1}{\color{MyRed2}\textbf{22}}& 4 & 4 & 0 & 0
&\scalebox{1.1}{\color{MyRed2}\textbf{19}}&\scalebox{1.1}{\color{MyRed2}\textbf{20}}&  7&  5 & 0 & 0
&\scalebox{1.1}{\color{MyRed2}\textbf{21}} & \scalebox{1.1}{\color{MyRed2}\textbf{17}} &6&6& 0 & 3
&\scalebox{1.1}{\color{MyRed2}\textbf{21}}&\scalebox{1.1}{\color{MyRed2}\textbf{19}}& 4 & 7 & 1 & 1 \\

\bottomrule[1.5pt]
\end{tabular}

}
\label{tab:main_result}
\vspace{-5pt}
\end{table*}

\section{Experiments}
\label{sec:exp}
We evaluate QUALS by retraining representative universal forecasters and assessing their zero-shot performance across large-scale benchmarks and systematic ablations.

\subsection{Experimental Setup}
\mypara{Baselines.}
\label{subsection:baselines}
We evaluate six models: the base and large versions of TimeMoE~\cite{shi2024timemoe} and MOIRAI~\cite{woo2024moirai}, and the small and base versions of Chronos\footnote{We employ the Chronos-Bolt release.}. Models without superscripts use official open-source checkpoints. Superscript B denotes BLAST-based versions~\footnote{https://huggingface.co/datasets/ZezhiShao/BLAST}, trained on an 81.1B corpus curated from a 321B raw corpus with potentially untraceable leakage. Superscript Q denotes models trained from scratch on QUALS, using an 81.89B candidate corpus constructed from a strictly leakage-free 239B raw corpus collected from GIFT-Eval Pretrain datasets~\footnote{https://huggingface.co/datasets/Salesforce/GiftEvalPretrain} and synthetic data~\cite{Chronos}. Training protocols otherwise follow the original implementations, with only batch size and gradient accumulation adjusted for TimeMoE-large to fit 4$\times$L40 GPUs. The QUALS length $L{=}4096$ is only the stored corpus-slice length; each model still uses its official context, masking, and rollout protocol. Notably, Stage~3 synchronization in QUALS enables convergence with fewer updates than the original settings.
Additional experiments on alternative corpus-construction strategies and domain-specific forecasting models are presented in Section~\ref{sec:baseline_comparison}.

\mypara{Benchmarks \& Data Leakage Declaration.}
To ensure fair and rigorous evaluation, we categorize our benchmarks based on their data leakage status (\ie whether test sets inadvertently appeared in official pre-training corpora)~\cite{shi2024timemoe, woo2024moirai, Chronos}:

\begin{itemize}[leftmargin=*, topsep=0pt, itemsep=2pt]
\item \textbf{Strictly Leakage-Free:} ARIES TEST\footnote{https://huggingface.co/spaces/Blisky-li/ARIES-TEST}~\cite{wang2025aries}, BLAST-configured GIFT-Eval~\cite{gifteval}.
\item \textbf{Leakage-Present (Declared):} We explicitly identify:
\begin{enumerate}[leftmargin=*, nosep]
\item \textit{LTSF}~\cite{li2025apt}\footnote{https://github.com/GestaltCogTeam/BasicTS}: ETTh1, ETTh2, ETTm1, ETTm2, and Weather; Weather is leaked in TimeMoE.
\item \textit{Full GIFT-Eval}\footnote{https://huggingface.co/spaces/Salesforce/GIFT-Eval}: partially leaked in TimeMoE, Chronos, their BLAST variants, and MOIRAI$^B$.
\item \textit{TSFM-Bench}\footnote{https://github.com/decisionintelligence/TSFM-Bench}~\cite{li2025tsfm}: our audit flags contamination on 13 of 21 evaluation datasets in TimeMoE, MOIRAI, Chronos, and their BLAST variants.

\item \textit{FEV-Bench}\footnote{https://huggingface.co/spaces/autogluon/fev-bench}~\cite{shchur2025fev}: partially leaked in TimeMoE, MOIRAI, their BLAST variants, and Chronos$^B$.
\end{enumerate}
These benchmarks are not leaked in QUALS-based retrained models.
\end{itemize}

\mypara{Metrics}
We report Mean Squared Error~(MSE) and Mean Absolute Error~(MAE) as the primary metrics in the main experiments. Under the regular setting, we report the mean and median values of MAE and MSE on ARIES TEST. For GIFT-EVAL, we adopt Mean Absolute Scaled Error~(MASE) and Continuous Ranked Probability Score~(CRPS). For TSFM-Bench, we likewise report MSE and MAE as skill scores relative to Seasonal Naive, following the FEV-Bench and GIFT-Eval convention due to space constraints.
For FEV-Bench, we report Scaled Quantile Loss~(SQL), Mean Absolute Scaled Error~(MASE), Weighted Quantile Loss~(WQL), and Weighted Absolute Percentage Error~(WAPE), all measured as skill scores relative to Seasonal Naive~\cite{makridakis2020m4}.

\subsection{Benchmark Comparison Experiment}
\mypara{Long-term Time Series Forecasting.} As presented in Table~\ref{tab:main_result}, models pre-trained on the QUALS corpus demonstrate consistent improvements of 3\% to 20\% (MSE/MAE) over their official open-source checkpoints. Compared to BLAST-balanced counterparts, QUALS achieves an additional 5–10\% MSE reduction in specific scenarios (\eg Chronos on ETTh1, TimeMoE on ETTh2 and ETTm1/2). Crucially, QUALS maintains a decisive lead even on the notoriously stable Weather dataset. Overall, QUALS secures the lowest errors across the most horizons, proving that corpus equilibrium translates directly into stronger long-term zero-shot forecasts.

Table~\ref{tab:benchmark_merged} summarizes zero-shot results on ARIES, GIFT-Eval-B/F, FEV, and TSFM-Bench under a unified layout.

\mypara{ARIES TEST.} On ARIES TEST, a synthetic benchmark with diverse temporal properties, baselines struggle with these patterns, while QUALS reduces the original MOIRAI MSE by over 60\%. This shows that balancing pattern distribution and learnability improves robust zero-shot forecasting.

\mypara{BLAST-configured GIFT-Eval.} To ensure a fair comparison free from the severe data leakage prevalent in massive pre-training corpora, we evaluate on the leakage-free BLAST-configured GIFT-Eval. Under this rigorous setting, QUALS models deliver relative error reductions of \textbf{7\% to 22\%} (MASE/CRPS) compared to their original versions, and yield an additional \textbf{1\% to 9\%} performance gain over the BLAST framework across all architectures.

\mypara{Full GIFT-Eval.} We further evaluate on the full 97-task GIFT-Eval.
In this setting, official corpora and BLAST models inherently possess an unfair memorization advantage due to explicit downstream data leakage. Astonishingly, despite using a strictly leakage-free and substantially smaller training budget, QUALS consistently outperforms all baselines. Most strikingly, it reduces the MASE of the leakage-advantaged TimeMoE models by \textbf{17\% to 23\%}, alongside consistent wins for Chronos and MOIRAI.

{\mypara{TSFM-Bench.} Table~\ref{tab:benchmark_merged} further reports TSFM-Bench results using MAE and MSE Skill Scores. Official pre-training corpora and BLAST partially leak TSFM-Bench test data, while QUALS remains strictly leakage-free. Nevertheless, QUALS-trained variants achieve the highest Skill Scores across all metrics and model families. Most strikingly, they rescue leakage-advantaged TimeMoE models that fall below Seasonal Naive and improve Skill Scores over original Chronos and MOIRAI checkpoints by \textbf{11\% to 50\%}, while also systematically eclipsing their BLAST counterparts.}

\mypara{FEV-Bench.} Table~\ref{tab:benchmark_merged} also reports FEV-Bench Skill Scores (vs.\ Seasonal Naive). Official corpora and BLAST leak FEV-Bench test data---e.g., MOIRAI has \textbf{28\%} leakage (28 of 100 configurations)---so BLAST can underperform originals by diluting this advantage. QUALS is retrained from leakage-free GIFT-Eval Pretrain with overlapping sources removed, yet still attains the highest Skill Scores, beating leakage-advantaged originals by 9\% to 18\% and their BLAST counterparts.


\begin{table*}[t]
\centering
\setlength{\abovecaptionskip}{0.cm}
\setlength{\belowcaptionskip}{0.cm}
\caption{
Cross-benchmark zero-shot results.
\textbf{GIFT-Eval-B}: BLAST-configured GIFT-Eval, retaining only leakage-free test data.
\textbf{GIFT-Eval-F}: full 97-task GIFT-Eval.
Models with superior performance are highlighted in {\color{MyRed2}\textbf{red}}.
}
\label{tab:ARIES}
\label{tab:gift_eval}
\label{tab:gift_eval_all}
\label{tab:fev}
\label{tab:tsfm_main}
\label{tab:benchmark_merged}
\setlength\tabcolsep{1.pt}
\renewcommand{\arraystretch}{0.85}
\resizebox{1\linewidth}{!}{%
\begin{tabular}{>{\centering\arraybackslash}p{2.0cm} c|ccc|ccc|ccc|ccc|ccc|ccc}
\toprule[1.5pt]
\multicolumn{2}{c|}{\textbf{Models}} &
\multicolumn{1}{l}{\textbf{TimeMoE}$_{l}^{\textbf{Q}}$} &
\multicolumn{1}{l}{\textbf{TimeMoE}$_{l}^\textbf{B}$} &
\multicolumn{1}{l|}{\textbf{TimeMoE}$_{l}$} &
\multicolumn{1}{l}{\textbf{TimeMoE}$_{b}^{\textbf{Q}}$} &
\multicolumn{1}{l}{\textbf{TimeMoE}$_{b}^\textbf{B}$} &
\multicolumn{1}{l|}{\textbf{TimeMoE}$_{b}$} &
\multicolumn{1}{l}{\textbf{MOIRAI}$_{l}^{\textbf{Q}}$} &
\multicolumn{1}{l}{\textbf{MOIRAI}$_{l}^\textbf{B}$} &
\multicolumn{1}{l|}{\textbf{MOIRAI}$_{l}$} &
\multicolumn{1}{l}{\textbf{MOIRAI}$_{b}^{\textbf{Q}}$} &
\multicolumn{1}{l}{\textbf{MOIRAI}$_{b}^\textbf{B}$} &
\multicolumn{1}{l|}{\textbf{MOIRAI}$_{b}$} &
\multicolumn{1}{l}{\textbf{Chronos}$_{b}^{\textbf{Q}}$} &
\multicolumn{1}{l}{\textbf{Chronos}$_{b}^\textbf{B}$} &
\multicolumn{1}{l|}{\textbf{Chronos}$_{b}$} &
\multicolumn{1}{l}{\textbf{Chronos}$_{s}^{\textbf{Q}}$} &
\multicolumn{1}{l}{\textbf{Chronos}$_{s}^{\textbf{B}}$} &
\multicolumn{1}{l}{\textbf{Chronos}$_{s}$} \\
\midrule
\multirow{2}{=}{\centering\textbf{ARIES-TEST}}
& \textbf{MAE}$\downarrow$
&{\color{MyRed2}\textbf{0.322}}&0.366&0.401
&{\color{MyRed2}\textbf{0.320}}&0.343&0.363
&{\color{MyRed2}\textbf{0.363}}&0.454&0.671
&{\color{MyRed2}\textbf{0.368}}&0.494&0.658
&{\color{MyRed2}\textbf{0.303}}&0.326&0.337
&{\color{MyRed2}\textbf{0.333}}&0.337&0.339 \\
& \textbf{MSE}$\downarrow$
&{\color{MyRed2}\textbf{0.353}}&0.412&0.424
&{\color{MyRed2}\textbf{0.347}}&0.415&0.411
&{\color{MyRed2}\textbf{0.412}}&0.522&1.012
&{\color{MyRed2}\textbf{0.435}}&0.583&2.193
&{\color{MyRed2}\textbf{0.313}}&0.338&0.349
&{\color{MyRed2}\textbf{0.338}}&0.349&0.352 \\
\midrule
\multirow{2}{=}{\centering\textbf{GIFT-Eval-B}}
& \textbf{MASE}$\downarrow$
&{\color{MyRed2}\textbf{0.677}}&0.742&0.872
&{\color{MyRed2}\textbf{0.734}}&0.760&0.888
&{\color{MyRed2}\textbf{0.729}}&0.740&0.816
&{\color{MyRed2}\textbf{0.745}}&0.759&0.812
&{\color{MyRed2}\textbf{0.659}}&0.682&0.724
&{\color{MyRed2}\textbf{0.663}}&0.684&0.736 \\
& \textbf{CRPS}$\downarrow$
&{\color{MyRed2}\textbf{0.557}}&0.593&0.702
&{\color{MyRed2}\textbf{0.593}}&0.616&0.706
&{\color{MyRed2}\textbf{0.576}}&0.580&0.642
&{\color{MyRed2}\textbf{0.585}}&0.614&0.641
&{\color{MyRed2}\textbf{0.450}}&0.468&0.489
&{\color{MyRed2}\textbf{0.453}}&0.470&0.485 \\
\midrule
\multirow{2}{=}{\centering\textbf{GIFT-Eval-F}}
& \textbf{MASE}$\downarrow$
&{\color{MyRed2}\textbf{0.846}}&0.852&1.110
&{\color{MyRed2}\textbf{0.881}}&0.896&1.067
&{\color{MyRed2}\textbf{0.858}}&0.880&0.875
&{\color{MyRed2}\textbf{0.888}}&0.908&0.901
&{\color{MyRed2}\textbf{0.798}}&0.804&0.808
&{\color{MyRed2}\textbf{0.803}}&0.809&0.822 \\
& \textbf{CRPS}$\downarrow$
&{\color{MyRed2}\textbf{0.723}}&0.728&0.882
&{\color{MyRed2}\textbf{0.759}}&0.766&0.906
&{\color{MyRed2}\textbf{0.580}}&0.598&0.599
&{\color{MyRed2}\textbf{0.608}}&0.623&0.610
&{\color{MyRed2}\textbf{0.568}}&0.574&0.574
&{\color{MyRed2}\textbf{0.571}}&0.574&0.577 \\
\midrule
\multirow{2}{=}{\centering\textbf{TSFM-Bench}}
& \textbf{MAE}$\uparrow$
&{\color{MyRed2}\textbf{0.139}}&0.111&-0.138
&{\color{MyRed2}\textbf{0.122}}&0.079&-0.167
&{\color{MyRed2}\textbf{0.154}}&0.153&0.103
&{\color{MyRed2}\textbf{0.162}}&0.151&0.117
&{\color{MyRed2}\textbf{0.215}}&0.050&0.188
&{\color{MyRed2}\textbf{0.199}}&0.175&0.174 \\
& \textbf{MSE}$\uparrow$
&{\color{MyRed2}\textbf{0.387}}&0.362&-0.017
&{\color{MyRed2}\textbf{0.371}}&0.321&-0.060
&{\color{MyRed2}\textbf{0.368}}&0.360&0.202
&{\color{MyRed2}\textbf{0.365}}&0.346&0.258
&{\color{MyRed2}\textbf{0.442}}&0.268&0.387
&{\color{MyRed2}\textbf{0.435}}&0.393&0.391 \\
\midrule
\multirow{4}{*}{\textbf{FEV}}
& \textbf{SQL}$\uparrow$
&{\color{MyRed2}\textbf{0.102}}&0.086&0.035
&{\color{MyRed2}\textbf{0.094}}&0.059&0.032
&{\color{MyRed2}\textbf{0.360}}&0.303&0.351
&{\color{MyRed2}\textbf{0.350}}&0.292&0.321
&{\color{MyRed2}\textbf{0.393}}&0.386&0.389
&{\color{MyRed2}\textbf{0.391}}&0.382&0.378 \\
& \textbf{MASE}$\uparrow$
&{\color{MyRed2}\textbf{0.110}}&0.095&0.044
&{\color{MyRed2}\textbf{0.102}}&0.067&0.040
&{\color{MyRed2}\textbf{0.226}}&0.151&0.216
&{\color{MyRed2}\textbf{0.214}}&0.140&0.181
&{\color{MyRed2}\textbf{0.275}}&0.266&0.265
&{\color{MyRed2}\textbf{0.275}}&0.260&0.256 \\
& \textbf{WQL}$\uparrow$
&{\color{MyRed2}\textbf{0.138}}&0.113&0.088
&{\color{MyRed2}\textbf{0.123}}&0.097&0.070
&0.398&0.352&{\color{MyRed2}\textbf{0.405}}
&{\color{MyRed2}\textbf{0.388}}&0.338&0.374
&{\color{MyRed2}\textbf{0.437}}&0.428&0.432
&{\color{MyRed2}\textbf{0.430}}&0.422&0.422 \\
& \textbf{WAPE}$\uparrow$
&{\color{MyRed2}\textbf{0.125}}&0.100&0.075
&{\color{MyRed2}\textbf{0.109}}&0.083&0.056
&0.247&0.192&{\color{MyRed2}\textbf{0.259}}
&{\color{MyRed2}\textbf{0.236}}&0.175&0.228
&{\color{MyRed2}\textbf{0.307}}&0.296&0.298
&{\color{MyRed2}\textbf{0.303}}&0.287&0.290 \\
\bottomrule[1.5pt]
\end{tabular}%
}
\end{table*}

\subsection{Baseline Comparison Experiment}\label{sec:baseline_comparison}

\subsubsection{\textbf{Corpus Strategies.}}\label{corpus_strategies}
We compare five corpus-construction alternatives by retraining ChronosBolt-Small and TimeMoE-Base on the same GIFT-Eval Pretrain pool: \textit{AvgDataset} balances sub-datasets, \textit{AvgDomain} balances domains, \textit{Catch22} uses catch22 features in the BLAST-style pipeline~\cite{lubba2019catch22}, \textit{BLAST}, and \textit{Online-GRPO} learns online weights on ChronosBolt-Small and transfers them.

As Table~\ref{tab:corpus_strategies} reports all benchmark results, QUALS consistently ranks first or second within each model family. On GIFT-Eval-B, ChronosBolt-Small improves by $\sim$10\% and $\sim$3\% over \textit{Original} and \textit{BLAST}; \textit{AvgDataset}, \textit{AvgDomain}, and \textit{Catch22} trail by larger margins. TimeMoE-Base leads on all LTSF datasets and delivers $\sim$3\% and $\sim$17\% GIFT-Eval-B gains over \textit{BLAST} and \textit{Original} with far fewer tokens. \textit{Online-GRPO} matches QUALS on a few FEV/TSFM metrics but not overall.
\providecommand{\best}[1]{{\color{MyRed2}\textbf{#1}}}
\providecommand{\second}[1]{{\color{MyBlue}\textbf{#1}}}

\begin{table*}[!t]
\centering
\setlength{\abovecaptionskip}{0.cm}
\setlength{\belowcaptionskip}{0.cm}
\setlength{\tabcolsep}{1.5pt}
\renewcommand{\arraystretch}{0.9}
\caption{
Performance comparison of different corpus construction strategies. 
\best{Red} and \second{Blue} indicate the best and second-best results within each model family, respectively.
}
\resizebox{1\linewidth}{!}{%
\begin{tabular}{r c|cccccccccc|cc|cc|cc|cc|cccc}
\toprule[1.5pt]
\multirow{2}{*}{} & \multirow{2}{*}{\textbf{Variant}} &
\multicolumn{2}{c}{\textbf{ETTh1}} &
\multicolumn{2}{c}{\textbf{ETTh2}} &
\multicolumn{2}{c}{\textbf{ETTm1}} &
\multicolumn{2}{c}{\textbf{ETTm2}} &
\multicolumn{2}{c|}{\textbf{Weather}} &
\multicolumn{2}{c|}{\textbf{ARIES-TEST}} &
\multicolumn{2}{c|}{\textbf{GIFT-Eval-B}} &
\multicolumn{2}{c|}{\textbf{GIFT-Eval-F}} &
\multicolumn{2}{c|}{\textbf{TSFM-Bench}} &
\multicolumn{4}{c}{\textbf{FEV}} \\
\cmidrule(lr){3-4}
\cmidrule(lr){5-6}
\cmidrule(lr){7-8}
\cmidrule(lr){9-10}
\cmidrule(lr){11-12}
\cmidrule(lr){13-14}
\cmidrule(lr){15-16}
\cmidrule(lr){17-18}
\cmidrule(lr){19-20}
\cmidrule(lr){21-24}
&
&
\textbf{MSE$\downarrow$} & \textbf{MAE$\downarrow$} &
\textbf{MSE$\downarrow$} & \textbf{MAE$\downarrow$} &
\textbf{MSE$\downarrow$} & \textbf{MAE$\downarrow$} &
\textbf{MSE$\downarrow$} & \textbf{MAE$\downarrow$} &
\textbf{MSE$\downarrow$} & \textbf{MAE$\downarrow$} &
\textbf{MSE$\downarrow$} & \textbf{MAE$\downarrow$} &
\textbf{MASE$\downarrow$} & \textbf{CRPS$\downarrow$} &
\textbf{MASE$\downarrow$} & \textbf{CRPS$\downarrow$} &
\textbf{MAE$\uparrow$} & \textbf{MSE$\uparrow$} &
\textbf{SQL$\uparrow$} & \textbf{MASE$\uparrow$} & \textbf{WQL$\uparrow$} & \textbf{WAPE$\uparrow$} \\
\midrule

\multirow{7}{*}{\rotatebox{90}{$\textbf{ChronosBolt}_s$}}
& {Original}
& .467 & .428
& .370 & .385
& .411 & .388
& .291 & .330
& .253 & .268
& .352 & .339
& .736 & .485
& .822 & .577
& .174 & .391
& .378 & .256 & .422 & .290 \\

& {AvgDataset}
& .414 & .412
& .355 & .381
& .386 & .391
& .329 & .351
& .261 & .282
& .794 & .596
& .757 & .522
& .886 & .629
& .126 & .283
& .379 & .255 & .427 & .293 \\

& {AvgDomain}
& .419 & .417
& .375 & .394
& \second{.382} & .389
& .303 & .335
& .251 & .268
& .729 & .566
& .755 & .524
& .882 & .628
& \best{.200} & \second{.429}
& .345 & .218 & .391 & .249 \\

& {Catch22}
& \second{.398} & .407
& \second{.350} & .380
& .679 & .486
& .327 & .358
& .275 & .290
& .798 & .596
& .793 & .545
& .939 & .670
& .116 & .341
& .332 & .198 & .372 & .226 \\

& {BLAST}
& .410 & .408
& .358 & .382
& .404 & .385
& \best{.282} & \best{.316}
& .249 & .268
& .349 & .337
& \second{.684} & .470
& \second{.809} & \second{.574}
& .175 & .393
& .382 & .260 & .422 & .287 \\

& {Online-GRPO}
& .403 & \second{.404}
& \best{.346} & \best{.371}
& .391 & \best{.380}
& .288 & \second{.321}
& \second{.247} & \second{.266}
& \best{.332} & \best{.305}
& .687 & \second{.461}
& .838 & .588
& .184 & .417
& \second{.383} & \second{.265} & \second{.428} & \second{.297} \\

\rowcolor{blue!15}\cellcolor{white!0}
& {QUALS}
& \best{.382} & \best{.399}
& \best{.346} & \second{.373}
& \best{.371} & \second{.381}
& \second{.287} & \second{.321}
& \best{.243} & \best{.263}
& \second{.338} & \second{.333}
& \best{.663} & \best{.453}
& \best{.803} & \best{.571}
& \second{.199} & \best{.435}
& \best{.391} & \best{.275} & \best{.430} & \best{.303} \\

\midrule

\multirow{7}{*}{\rotatebox{90}{$\textbf{TimeMoE}_{b}$}}

& {Original}
& .400 & .424
& .366 & .404
& \second{.394} & .415
& .317 & .365
& .274 & .297
& .411 & .363
& .888 & .706
& 1.067 & .906
& -0.167& -0.060
& .032 & .040 & .070 & .056 \\

& {AvgDataset}
& .432 & .445
& .361 & .391
& .443 & .423
& .314 & .353
& .263 & .301
& .731 & .604
& 1.067 & .819
& 1.031 & .873
& .058 & .319
& .002 & .011 & .035 & .021 \\

& {AvgDomain}
& .407 & .423
& .356 & \second{.388}
& .428 & .417
& .303 & .346
& .257 & .292
& .728 & .604
& 1.056 & .811
& 1.021 & .866
& .085 & .343
& -0.022 & -0.013 & -0.000 & -0.015 \\

& {Catch22}
& \second{.393} & .413
& .362 & .398
& .688 & .535
& .313 & .361
& .271 & .307
& .757 & .613
& 1.066 & .819
& 1.086 & .937
& .025 & .280
& -0.021 & -0.011 & .004 & -0.011 \\

& {BLAST}
& .399 & .412
& .356 & .391
& \second{.394} & .399
& \second{.274} & \second{.326}
& \best{.244} & \best{.278}
& .415 & \second{.343}
& \second{.760} & \second{.616}
& \second{.896} & \second{.766}
& .079 & .321
& .059 & .067 & .097 & .083 \\

& {Online-GRPO}
& .394 & \second{.407}
& \second{.352} & .389
& .397 & \second{.391}
& .276 & \second{.326}
& .255 & .290
& \second{.390} & .354
& .841 & .686
& .949 & .814
& \best{.123} & \best{.374}
& \best{.095} & \best{.103} & \second{.118} & \second{.105} \\

\rowcolor{blue!15}\cellcolor{white!0}
& 
{QUALS}
& \best{.383} & \best{.405}
& \best{.338} & \best{.384}
& \best{.354} & \best{.378}
& \best{.251} & \best{.316}
& \second{.248} & \second{.282}
& \best{.347} & \best{.320}
& \best{.734} & \best{.593}
& \best{.881} & \best{.759}
& \second{.122} & \second{.371}
& \second{.094} & \second{.102} & \best{.123} & \best{.109} \\

\bottomrule[1.5pt]
\end{tabular}%
}
\label{tab:corpus_strategies}
\end{table*}

\subsubsection{\textbf{Domain-Specific Models.}}\label{domain_specific_models}
We compare QUALS-pretrained universal forecasters with domain-specific full-shot models on ETTh1, ETTh2, ETTm1, ETTm2, and Weather. As Table~\ref{tab:domain_specific_main} reports, QUALS variants rank first or second on most splits and outperform domain-specific models on average.
\begin{table*}[!t]
\setlength{\abovecaptionskip}{0.cm}
\caption{
Comparison of zero-shot models pretrained on QUALS and domain-specific models. 
{\color{MyRed2}\textbf{Red}} and {\color{MyBlue}Blue} indicate the best and second-best results, respectively.
}
\setlength\tabcolsep{1.5pt}
\renewcommand{\arraystretch}{0.9}
\resizebox{1\linewidth}{!}{%
\begin{tabular}{rr|cc|cc|cc|cc|cc|cc|cc|cc|cc|cc|cc|cc|cc|cc}
\toprule[1.5pt]
\multicolumn{2}{l}{} & \multicolumn{12}{c}{\textbf{Pre-training on QUALS}} & \multicolumn{16}{c}{\textbf{Domain-specific Models}}\\
\cmidrule(r){3-14} \cmidrule(r){15-30}
\multicolumn{2}{r}{\textbf{Models}} & \multicolumn{2}{|c}{\textbf{TimeMoE}$_{b}^{\textbf{Q}}$} & \multicolumn{2}{c|}{\textbf{TimeMoE}$_{l}^{\textbf{Q}}$} & \multicolumn{2}{c|}{\textbf{MOIRAI}$_{b}^{\textbf{Q}}$} & \multicolumn{2}{c|}{\textbf{MOIRAI}$_{l}^{\textbf{Q}}$} & \multicolumn{2}{c|}{\textbf{Chronos}$_{b}^{\textbf{Q}}$} & \multicolumn{2}{c|}{\textbf{Chronos}$_{s}^{\textbf{Q}}$} & \multicolumn{2}{c|}{\begin{tabular}[c]{@{}c@{}}\textbf{TQNet}\\2025~\cite{lin2025temporal}\end{tabular}} & \multicolumn{2}{c|}{\begin{tabular}[c]{@{}c@{}}\textbf{OLinear}\\2025~\cite{yue2026olinear}\end{tabular}} & \multicolumn{2}{c|}{\begin{tabular}[c]{@{}c@{}}\textbf{TimeMixer}\\2024~\cite{wangtimemixer}\end{tabular}} & \multicolumn{2}{c|}{\begin{tabular}[c]{@{}c@{}}\textbf{iTransformer}\\2024~\cite{liuitransformer}\end{tabular}} & \multicolumn{2}{c|}{\begin{tabular}[c]{@{}c@{}}\textbf{TimesNet}\\2023~\cite{DBLP:conf/iclr/WuHLZ0L23}\end{tabular}} & \multicolumn{2}{c|}{\begin{tabular}[c]{@{}c@{}}\textbf{PatchTST}\\2023~\cite{nietime}\end{tabular}} & \multicolumn{2}{c|}{\begin{tabular}[c]{@{}c@{}}\textbf{TiDE}\\2023~\cite{daslong}\end{tabular}}& \multicolumn{2}{c}{\begin{tabular}[c]{@{}c@{}}\textbf{DLinear}\\2023~\cite{zeng2023transformers}\end{tabular}} \\
\cmidrule(r){3-4} \cmidrule(r){5-6} \cmidrule(r){7-8} \cmidrule(r){9-10} \cmidrule(r){11-12} \cmidrule(r){13-14} \cmidrule(r){15-16} \cmidrule(r){17-18} \cmidrule(r){19-20} \cmidrule(r){21-22} \cmidrule(r){23-24} \cmidrule(r){25-26} \cmidrule(r){27-28} \cmidrule(r){29-30}
\multicolumn{2}{r|}{\textbf{Metrics}} & \textbf{MSE} & \textbf{MAE} & \textbf{MSE} & \textbf{MAE} & \textbf{MSE} & \textbf{MAE} & \textbf{MSE} & \textbf{MAE} & \textbf{MSE} & \textbf{MAE} & \textbf{MSE} & \textbf{MAE} & \textbf{MSE} & \textbf{MAE} & \textbf{MSE} & \textbf{MAE} & \textbf{MSE} & \textbf{MAE} & \textbf{MSE} & \textbf{MAE} & \textbf{MSE} & \textbf{MAE} & \textbf{MSE} & \textbf{MAE} & \textbf{MSE} & \textbf{MAE} & \textbf{MSE} & \textbf{MAE} \\
\midrule
\multicolumn{2}{r|}{\rotatebox{0}{ETTh1}}
&\color{MyBlue}\textbf{.383}&.405&.389&.408&.407&.408&.401&.404&.391&\color{MyBlue}\textbf{.401}&\color{MyRed2}\textbf{.382}&\color{MyRed2}\textbf{.399}&.437&.433&.444&.446&.448&.442&.454&.447&.454&.450&.468&.454&.540&.507&.455&.451\\
\multicolumn{2}{r|}{\rotatebox{0}{ETTh2}}
&\color{MyRed2}\textbf{.338}&.384&.345&.387&.344&\color{MyRed2}\textbf{.369}&\color{MyBlue}\textbf{.341}&\color{MyBlue}\textbf{.370}&.349&.374&.346&.373&.390&.417&.393&.420&.364&.395&.383&.406&.414&.496&.386&.406&.611&.549&.558&.515\\
\multicolumn{2}{r|}{\rotatebox{0}{ETTm1}}
&.354&.378&\color{MyRed2}\textbf{.346}&.375&.377&.378&.386&.377&.390&.379&.371&.381&.355&\color{MyRed2}\textbf{.371}&\color{MyBlue}\textbf{.350}&\color{MyBlue}\textbf{.374}&.381&.395&.407&.409&.400&.405&.387&.400&.419&.419&.403&.406\\
\multicolumn{2}{r|}{\rotatebox{0}{ETTm2}}
&\color{MyBlue}\textbf{.251}&.316&\color{MyRed2}\textbf{.243}&\color{MyBlue}\textbf{.310}&.274&.317&.259&\color{MyRed2}\textbf{.303}&.275&.313&.287&.321&.259&.312&.263&.314&.275&.323&.288&.332&.291&.332&.280&.326&.358&.403&.350&.400\\
\multicolumn{2}{r|}{\rotatebox{0}{Weather}}
&.248&.282&.248&.280&.258&.272&.247&.265&.240&.259&.243&.263&\color{MyRed2}\textbf{.221}&\color{MyRed2}\textbf{.249}&\color{MyBlue}\textbf{.226}&\color{MyBlue}\textbf{.250}&.240&.271&.257&.278&.258&.286&.258&.280&.270&.320&.265&.316\\
\midrule
\rowc\rowcolor{blue!15}
\multicolumn{2}{c|}{\scalebox{1.1}{\textbf{Average}}}
&{\color{MyBlue}\textbf{.315}}&.353&{\color{MyRed2}\textbf{.314}}&.352&.332&.349&.327&{\color{MyRed2}\textbf{.344}}&.329&{\color{MyBlue}\textbf{.345}}&.326&.347&.332&.356&.335&.361&.342&.365&.358&.374&.363&.394&.356&.373&.440&.440&.406&.418\\
\bottomrule[1.5pt]
\end{tabular}}
\label{tab:domain_specific_main}
\end{table*}

\begin{table*}[t]
\setlength{\abovecaptionskip}{0.cm}
\setlength{\belowcaptionskip}{0.cm}
\centering

\setlength{\tabcolsep}{1pt}
\renewcommand{\arraystretch}{0.85}
\caption{Ablation study of QUALS. Variants with superior performance are highlighted in {\color{MyRed2}\textbf{red}}.
}
\label{tab:ablation_transposed_quals_321_best}

\resizebox{0.98\textwidth}{!}{%
\begin{tabular}{c c ccccc ccccc ccccc ccccc}
\toprule[1.5pt]
\multicolumn{2}{c}{} &
\multicolumn{5}{c}{\textbf{QUALS}} &
\multicolumn{5}{c}{\textbf{w/o Stage3} (Candidate Corpus)} &
\multicolumn{5}{c}{\textbf{w/o Stage2} (Codebook Sample)} &
\multicolumn{5}{c}{\textbf{w/o Stage1} (BLAST)} \\
\cmidrule(lr){3-7}\cmidrule(lr){8-12}\cmidrule(lr){13-17}\cmidrule(lr){18-22}
\textbf{Model} & \textbf{Metric} &
ETTh1 & ETTh2 & ETTm1 & ETTm2 & Weather &
ETTh1 & ETTh2 & ETTm1 & ETTm2 & Weather &
ETTh1 & ETTh2 & ETTm1 & ETTm2 & Weather &
ETTh1 & ETTh2 & ETTm1 & ETTm2 & Weather \\
\midrule

\multirow{2}{*}{\textbf{ChronosBolt}$_s$} & MSE$\downarrow$ &
{\color{MyRed2}\textbf{.382}} & {\color{MyRed2}\textbf{.346}} & {\color{MyRed2}\textbf{.371}} & .287 & {\color{MyRed2}\textbf{.243}} &
.397 & .354 & .386 & .285 & {\color{MyRed2}\textbf{.243}} &
.419 & .362 & .405 & {\color{MyRed2}\textbf{.278}} & .247 &
.411 & .358 & .404 & .282 & .249 \\
& MAE$\downarrow$ &
{\color{MyRed2}\textbf{.399}} & {\color{MyRed2}\textbf{.373}} & .381 & .321 & .263 &
.402 & .381 & {\color{MyRed2}\textbf{.379}} & .314 & {\color{MyRed2}\textbf{.262}} &
.412 & .384 & .384 & {\color{MyRed2}\textbf{.313}} & .264 &
.409 & .382 & .382 & .316 & .268 \\
\midrule

\multirow{2}{*}{\textbf{ChronosBolt}$_b$} & MSE$\downarrow$ &
{\color{MyRed2}\textbf{.391}} & {\color{MyRed2}\textbf{.349}} & {\color{MyRed2}\textbf{.390}} & {\color{MyRed2}\textbf{.275}} & {\color{MyRed2}\textbf{.240}} &
.393 & .352 & .401 & .280 & .248 &
.404 & .350 & .397 & .282 & .259 &
.409 & .355 & .390 & .277 & .244 \\
& MAE$\downarrow$ &
{\color{MyRed2}\textbf{.401}} & {\color{MyRed2}\textbf{.374}} & .379 & {\color{MyRed2}\textbf{.313}} & {\color{MyRed2}\textbf{.259}} &
.403 & .380 & .381 & .321 & .269 &
.410 & .381 & .384 & .322 & .281 &
.409 & .377 & {\color{MyRed2}\textbf{.375}} & {\color{MyRed2}\textbf{.313}} & .263 \\
\midrule

\multirow{2}{*}{\textbf{TimeMoE}$_b$} & MSE$\downarrow$ &
{\color{MyRed2}\textbf{.383}} & {\color{MyRed2}\textbf{.338}} & {\color{MyRed2}\textbf{.354}} & {\color{MyRed2}\textbf{.251}} & .248 &
.398 & .341 & .360 & .254 & {\color{MyRed2}\textbf{.242}} &
.405 & {\color{MyRed2}\textbf{.338}} & .363 & .269 & .255 &
.399 & .356 & .394 & .274 & .244 \\
& MAE$\downarrow$ &
{\color{MyRed2}\textbf{.405}} & .384 & {\color{MyRed2}\textbf{.378}} & {\color{MyRed2}\textbf{.316}} & .282 &
.407 & {\color{MyRed2}\textbf{.382}} & .382 & .318 & {\color{MyRed2}\textbf{.274}} &
.416 & .383 & .381 & .319 & .288 &
.412 & .391 & .399 & .326 & .278 \\
\bottomrule[1.5pt]
\end{tabular}%
}
\end{table*}

\begin{table}[h]
\setlength{\abovecaptionskip}{0.cm}
\setlength{\belowcaptionskip}{0.cm}
\renewcommand\arraystretch{0.7}
\scriptsize
\centering
\caption{QUALS vs. BLAST: scale and efficiency.}
\label{tab:quals_blast_efficiency}
\scalebox{1.2}{%
\begin{tabular}{c|c|c}
\toprule
\textbf{Metric} & \textbf{QUALS} & \textbf{BLAST} \\
\midrule
Raw Corpus Scale      &239 B  & 321 B \\
Curated Corpus Scale      &81.89 B  & 81.10 B  \\
\midrule
Representation Time          & 20 GPU-hours &  336 CPU-hours\\
Sampling Time                 & 12 CPU-hours  & 18 CPU-hours \\
Rebalancing Time         & 7 GPU-hours &  -\\
\bottomrule
\end{tabular}%
}
\vspace{-5pt}
\end{table}

\begin{table}[t]
\setlength{\abovecaptionskip}{0.cm}
\setlength{\belowcaptionskip}{0.cm}
\renewcommand\arraystretch{0.7}
\footnotesize
\centering
\caption{Zero-Shot performance (MSE / MAE) across 5 benchmarks in Table~\ref{tab:main_result} and their training efficiency (tokens). 
Bottom value represents the pre-training token budget (Billions, B). 
"-" indicates the pretraining budgets is not disclosed.}
\label{tab:performance_efficiency}
\begin{tabular}{l|c|c|c}
\toprule[1.2pt]
\textbf{Model} & \textbf{QUALS (Ours)} & \textbf{BLAST} & \textbf{Official Baseline} \\
\midrule

\textbf{ChronosBolt-Small} & 
\color{MyRed2}\textbf{\splitcell{0.326 / 0.347}{61.86 B}} & 
\splitcell{0.341 / 0.352}{104.85 B} & 
\splitcell{0.358 / 0.360}{-} 
\\ \midrule

\textbf{ChronosBolt-Base} & 
\color{MyRed2}\textbf{\splitcell{0.329 / 0.345}{125.82 B}} & 
\splitcell{0.335 / 0.347}{167.77 B} & 
\splitcell{0.351 / 0.351}{-} 
\\ \midrule

\textbf{MOIRAI-Base} & 
\color{MyRed2}\textbf{\splitcell{0.332 / 0.349}{35.360 B}} & 
\splitcell{0.338 / 0.353}{52.400 B} & 
\splitcell{0.357 / 0.361}{52.400 B} 
\\ \midrule

\textbf{MOIRAI-Large} & 
\color{MyRed2}\textbf{\splitcell{0.327 / 0.344}{38.640 B}} & 
\splitcell{0.335 / 0.350}{52.400 B} & 
\splitcell{0.381 / 0.393}{52.400 B} 
\\ \midrule

\textbf{TimeMoE-Base} & 
\color{MyRed2}\textbf{\splitcell{0.315 / 0.353}{26.20 B}} & 
\splitcell{0.333 / 0.361}{78.64 B} & 
\splitcell{0.350 / 0.381}{419.43 B} 
\\ \midrule

\textbf{TimeMoE-Large} & 
\color{MyRed2}\textbf{\splitcell{0.314 / 0.352}{99.61 B}} & 
\splitcell{0.330 / 0.360}{209.71 B} & 
\splitcell{0.352 / 0.380}{419.43 B} 
\\ \bottomrule[1.2pt]

\end{tabular}
\end{table}

\subsection{Efficiency Analysis}\label{Efficiency}

In this section, we first compare QUALS and BLAST in data scale and processing efficiency. We then analyze pre-training data efficiency across QUALS, BLAST, and the original baselines, focusing on the trade-off between training budget and forecasting performance.

\mypara{Data Scale.} Early paradigms assume that scaling inherently improves performance, leading BLAST to aggregate a $\sim$321B corpus (Table~\ref{tab:quals_blast_efficiency}). To prevent data leakage, QUALS restricts its raw corpus to $\sim$239B points. Moreover, expanded corpora can be processed via VQ-SwinTTS inference, avoiding costly statistical features.

\mypara{Data Processing Efficiency.} Building upon the computational accounting in Section~\ref{pipeline}, we compare the end-to-end efficiency of QUALS and BLAST, as summarized in Table~\ref{tab:quals_blast_efficiency}. The fundamental advantage of QUALS lies in its paradigm shift from CPU-bound mathematical heuristics to a highly scalable, GPU-accelerated pipeline.

Specifically, BLAST's reliance on statistical feature extraction incurs a prohibitive 336 CPU-hours on server-grade hardware and scales poorly to massive datasets. Furthermore, its high-complexity UMAP dimensionality reduction forces it to down-sample and train on merely 1\% of the raw corpus, causing irreversible information loss. In contrast, QUALS bypasses these bottlenecks entirely. By replacing CPU-intensive feature engineering with rapid VQ-SwinTTS inference and substituting UMAP with efficient PCA and ECDF, QUALS processes the full 239B corpus in a strictly lossless manner. Even with the introduction of the novel learnability synchronization stage (Stage~3), the additional overhead remains remarkably low ($\sim$7 GPU-hours). Ultimately, QUALS reduces the end-to-end preparation timeframe from weeks to under two days while eliminating the compromises introduced by down-sampling, thereby establishing a robust foundation for billion-scale pre-training.

\mypara{Pre-training Budget Efficiency.}
Table~\ref{tab:performance_efficiency} challenges the conventional scaling law for universal time series forecasting by demonstrating that {QUALS} achieves superior zero-shot accuracy with drastically reduced data volume. Most notably, for {TimeMoE-Base}, QUALS achieves a significantly lower MSE~(0.315 vs.\ 0.350) using only \textbf{26.20B} tokens---a \textbf{16$\times$ reduction} compared to the official 419.43B budget. This implies that over 90\% of the training tokens constitute redundant patterns that yield diminishing returns. This efficiency gain is consistent across architectures: {ChronosBolt-Small} reduces the token budget by \textbf{40\%} compared to BLAST while improving MSE by 4.4\%, and {MOIRAI-Large} outperforms its official checkpoint and the BLAST counterpart with 26\% fewer tokens. These results confirm that optimizing corpus equilibrium with respect to pattern distribution and learnability, rather than merely increasing scale, yields a far more efficient pre-training paradigm.

\begin{figure}[h]
  \centering
\setlength{\abovecaptionskip}{0.cm}
\setlength{\belowcaptionskip}{0.cm}
  \includegraphics[width=0.9\linewidth]{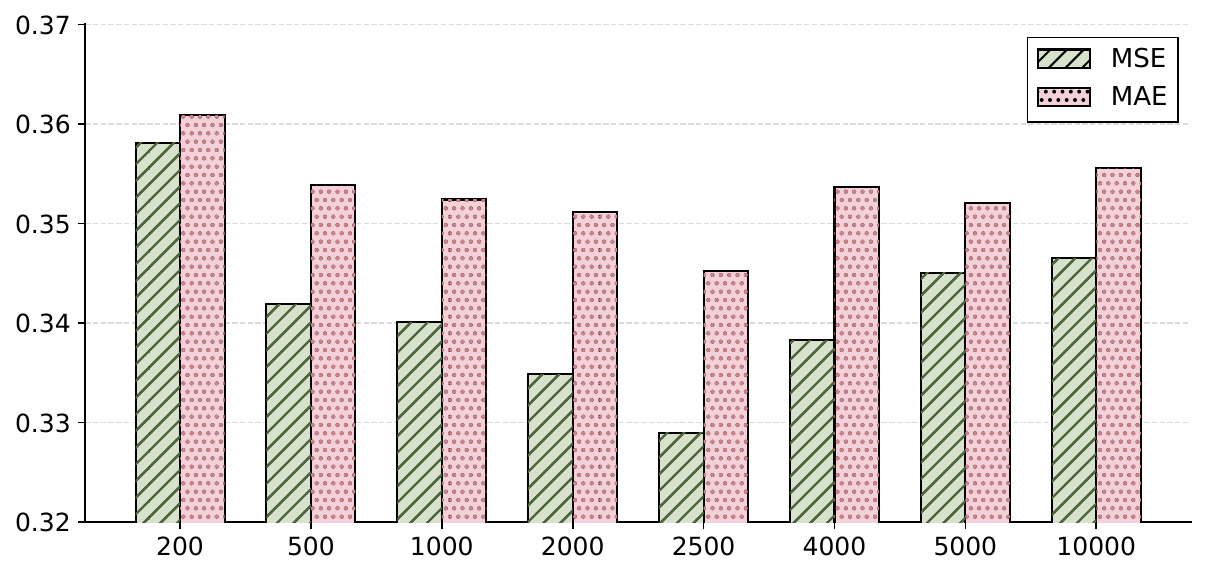}
  \caption{Impact of Stage 2 bin granularity of ECDF.}
  \label{fig:xr}
\end{figure}

\subsection{Ablation \& Hyper-parameter Studies}
Table~\ref{tab:ablation_transposed_quals_321_best} presents a stepwise ablation study validate the role of each QUALS component. Removing Stage 3 (learnability synchronization) consistently increases forecasting errors, proving that mere distribution balancing is insufficient.
Further removing Stage 2 (pattern binning) and relying instead on naive codebook sampling exacerbates the degradation, confirming that our spectral discretization is vital to prevent long-tail motif loss. Finally, replacing Stage 1's VQ-SwinTTS with BLAST's statistical heuristics yields the worst performance. This shows that deep semantic quantization captures temporal morphologies far more effectively than rigid metrics.

Figure~\ref{fig:xr} illustrates the sensitivity to the number of pattern bins ($M$) in Stage 2, evaluated under a fixed total training volume. We observe a distinct U-shaped trade-off: a small $M$ (\eg 200) causes under-fitting by collapsing diverse motifs into coarse bins, whereas excessive bins ($>5000$) lead to over-fragmentation and data sparsity. Optimal performance occurs between 2,000 and 4,000 bins, where QUALS successfully hits the "sweet spot" balancing high-resolution pattern coverage and sufficient sampling density.

\subsection{Deeper Analysis \& Visualization}

In this section, we analyze QUALS to validate the role of each stage in its three-stage framework. Our goal is to explain \emph{why} QUALS works, rather than merely reporting performance gains. To this end, we examine the method from pattern disentanglement, pattern organization, learnability synchronization, and forecasting behavior, thereby connecting the three stages of QUALS to its robustness in complex time-series scenarios.

\subsubsection{\textbf{Codebook Disentanglement}}\label{pdd}

This subsection validates {Stage 1} by examining whether VQ-SwinTTS learns semantically separated codebook. We verify this through similarity distributions and patch visualizations across different codebook indices.

\begin{figure}[!ht]
  \centering
\setlength{\abovecaptionskip}{0.cm}
\setlength{\belowcaptionskip}{0.cm}
  \includegraphics[width=0.9\linewidth]{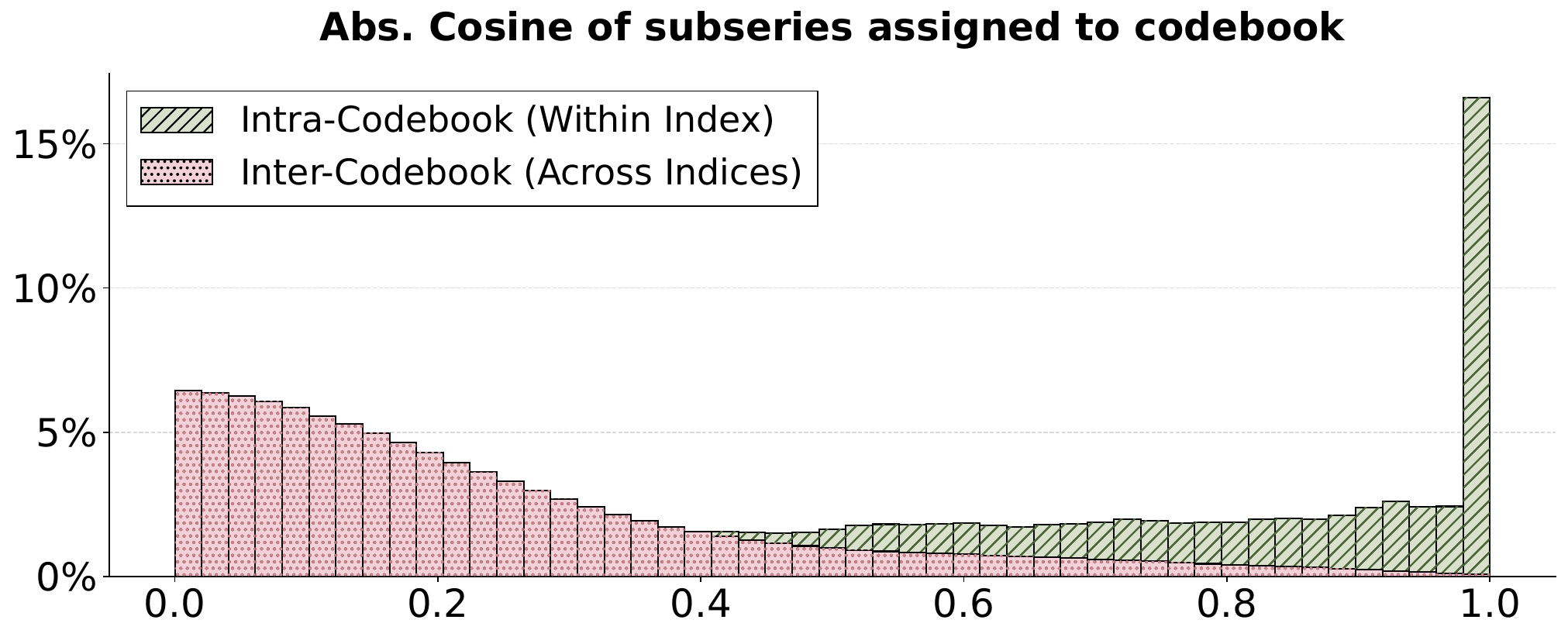}
  \caption{Distribution of Abs. cosine for inter- and intra-codebook subseries.}
  \label{fig:sim_dist}
\end{figure}

\begin{figure}[!ht]
  \centering
    \setlength{\abovecaptionskip}{0.0cm}
    \setlength{\belowcaptionskip}{0.cm}
  \includegraphics[width=0.9\linewidth]{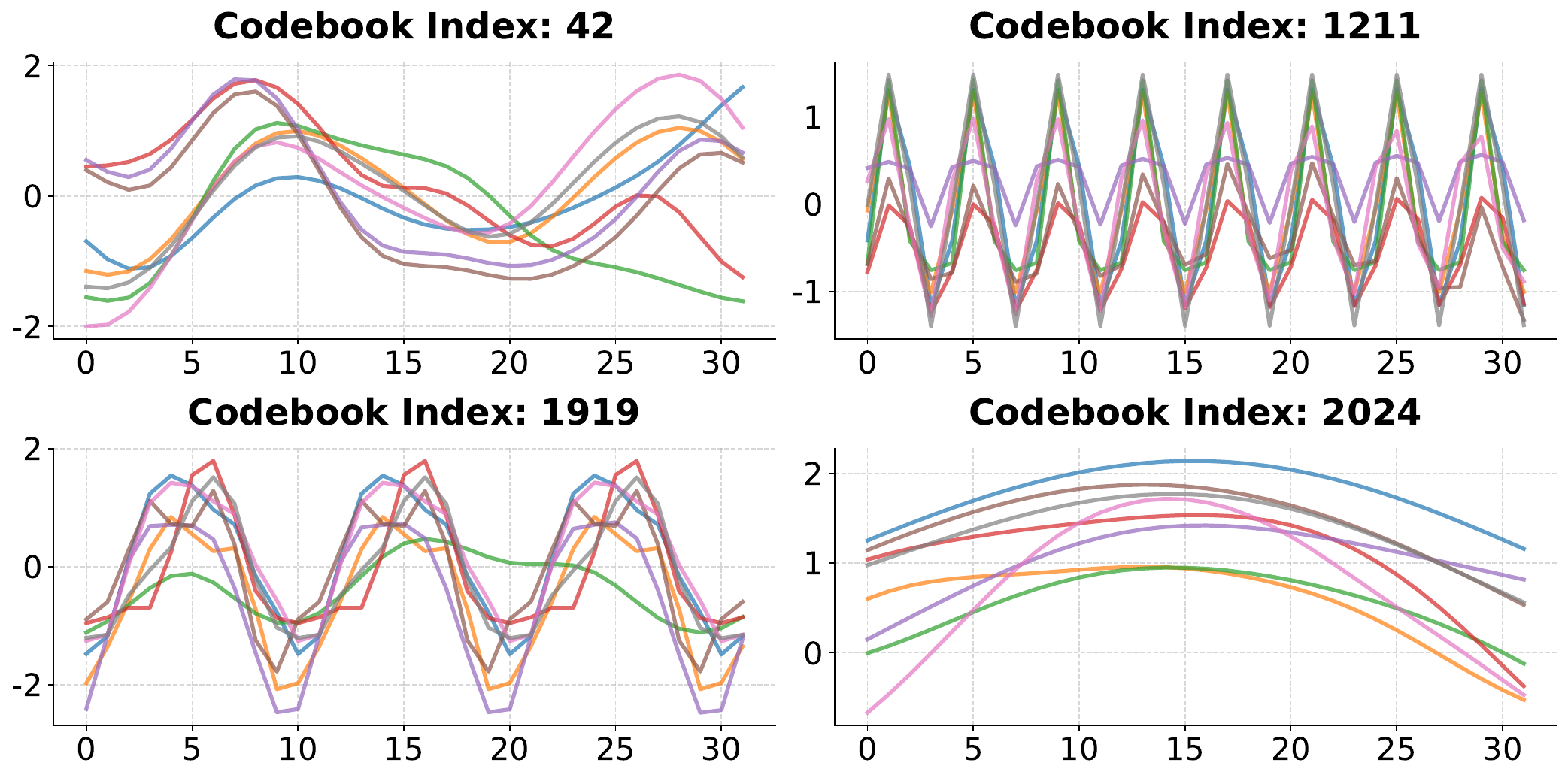}
  \caption{Visualization of learned patch-level pattern.}
  \label{fig:sim_dist2}
\end{figure}

Using the ARIES Synth dataset~\cite{wang2025aries}, a synthetic dataset with rich pattern diversity, we compare the absolute cosine similarity distributions of intra-code and inter-code subseries. Figure~\ref{fig:sim_dist} shows that subseries assigned to the same codebook index are much more similar than those assigned to different indices, indicating clear within-code consistency and cross-code separation.  Figure~\ref{fig:sim_dist2} further shows that different codebook indices capture distinct local motifs, while preserving strong regularity within each index. These results confirm that Stage~1 organizes patches into clearly separated patch-level semantics before the higher-level binning stage.

\subsubsection{\textbf{Pattern Space Organization}} This subsection validates {Stage 2} by showing how QUALS transfers information from a codebook space with directionally concentrated spectral energy to a more uniform space. We then examine semantic preservation after merging and uneven 
pattern distributions of existing datasets.

\begin{figure}[!t]
  \centering
\setlength{\abovecaptionskip}{0.cm}
\setlength{\belowcaptionskip}{0.cm}
  \includegraphics[width=0.9\linewidth]{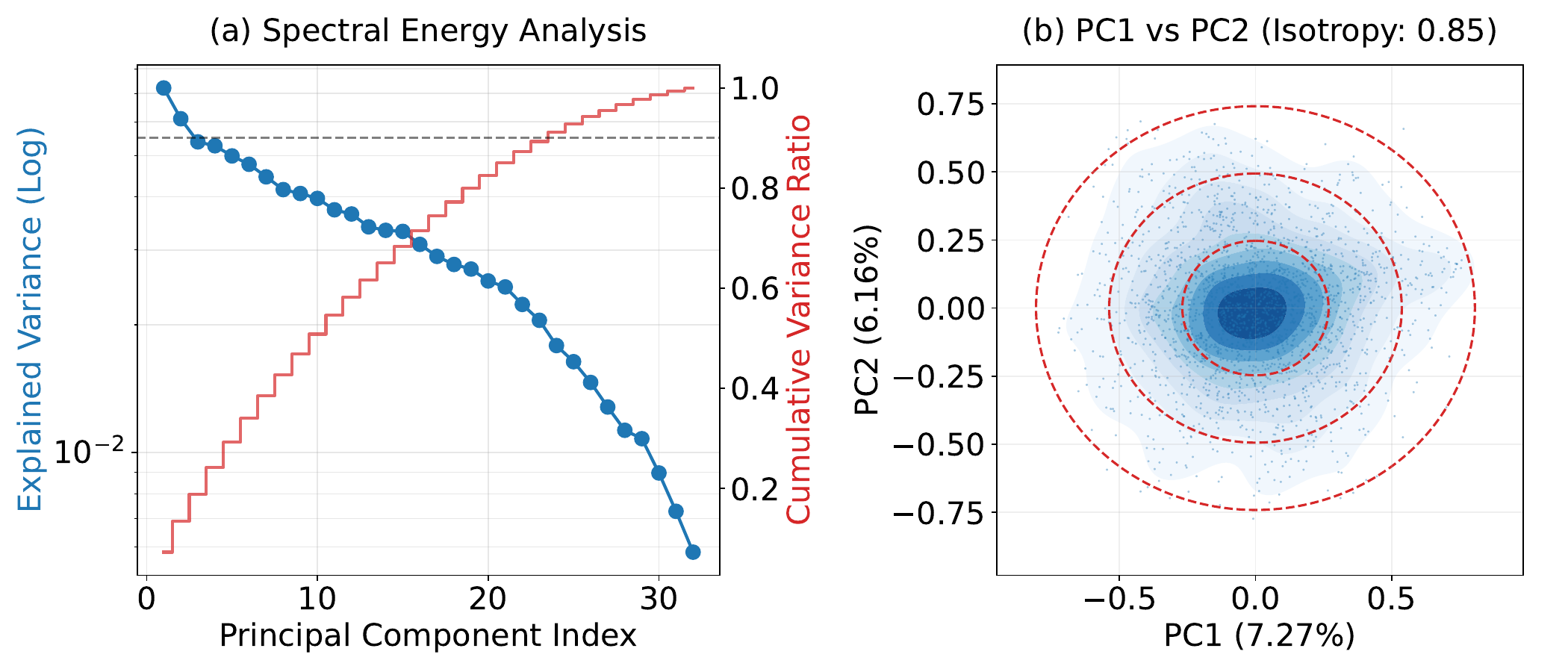}
  \caption{Spectral and geometric analysis of the codebook.}
  \label{fig:codebook_analysis}
\end{figure}

\mypara{Spectral Distribution.}
Analyzing the principal components of the learned codebook $\mathcal{C}$ reveals a slow, nearly linear eigenvalue decay (Figure~\ref{fig:codebook_analysis}(a)). In particular, 28 of 32 dimensions are required to capture 90\% of the spectral energy, indicating that semantic information spreads across nearly all dimensions while remaining concentrated along different directions in a non-uniform manner. Figure~\ref{fig:codebook_analysis}(b) further shows a highly isotropic 2D projection on the primary PCA axes, where the variance ratio $PC_2/PC_1$ reaches 0.85. Together, these results suggest that the codebook retains substantial high-dimensional information before Stage~2 merging.

\begin{figure}[!t]
  \centering
    \setlength{\abovecaptionskip}{0.cm}
    \setlength{\belowcaptionskip}{0.cm}
  \includegraphics[width=0.9\linewidth]{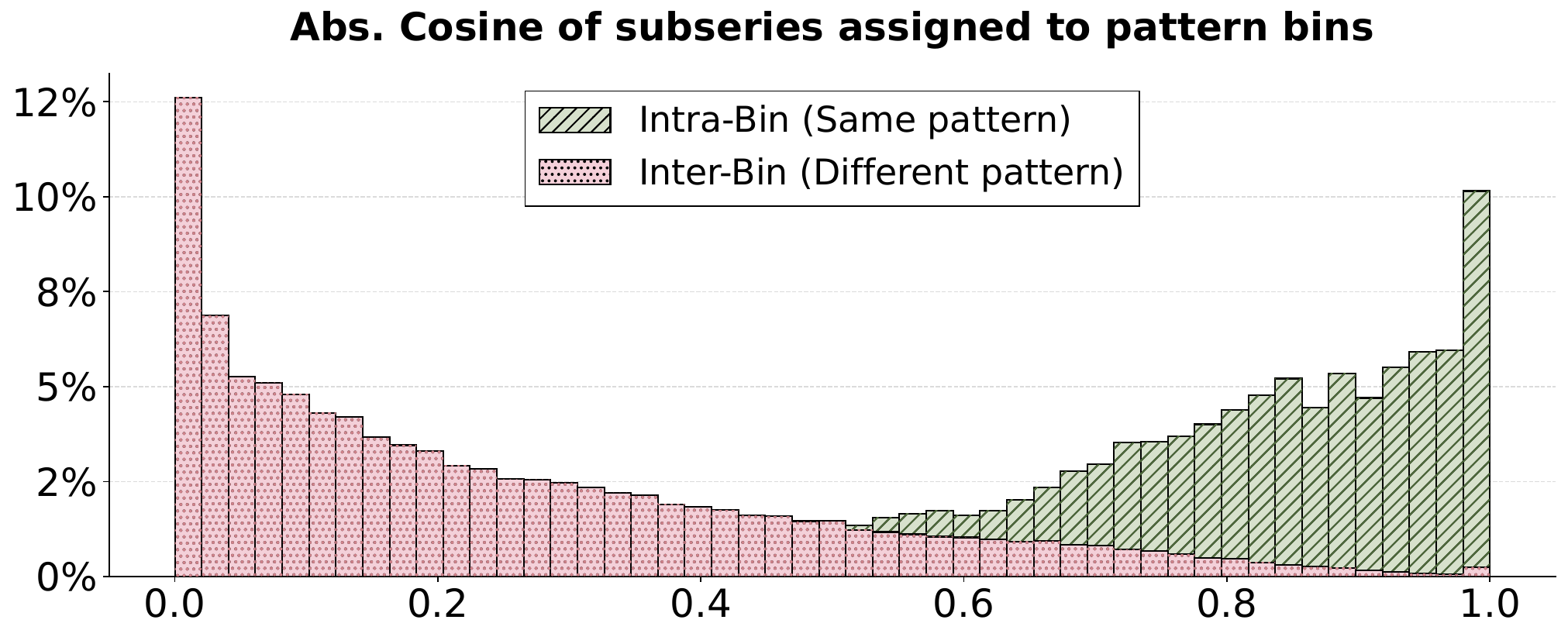}
  \caption{Distribution of Abs. cosine for subseries from inter- and intra- pattern bins.}
  \label{fig:sim_pattern}
\end{figure}

\mypara{Semantic Preservation.}
To evaluate whether Stage~2 preserves semantic information after merging patch-level codebook indices into series-level Pattern Bins ($\mathcal{U}$), we compare the cosine similarity distributions of subseries assigned to pattern bins in Figure~\ref{fig:sim_pattern} with the codebook-level results in Figure~\ref{fig:sim_dist}, and find a modest increase in overlap, indicating slight semantic drift but largely preserved separation at the bin level.

\begin{figure}[!b]
  \centering
    \setlength{\abovecaptionskip}{0.cm}´
    \setlength{\belowcaptionskip}{0.cm}
  \includegraphics[width=0.9\linewidth]{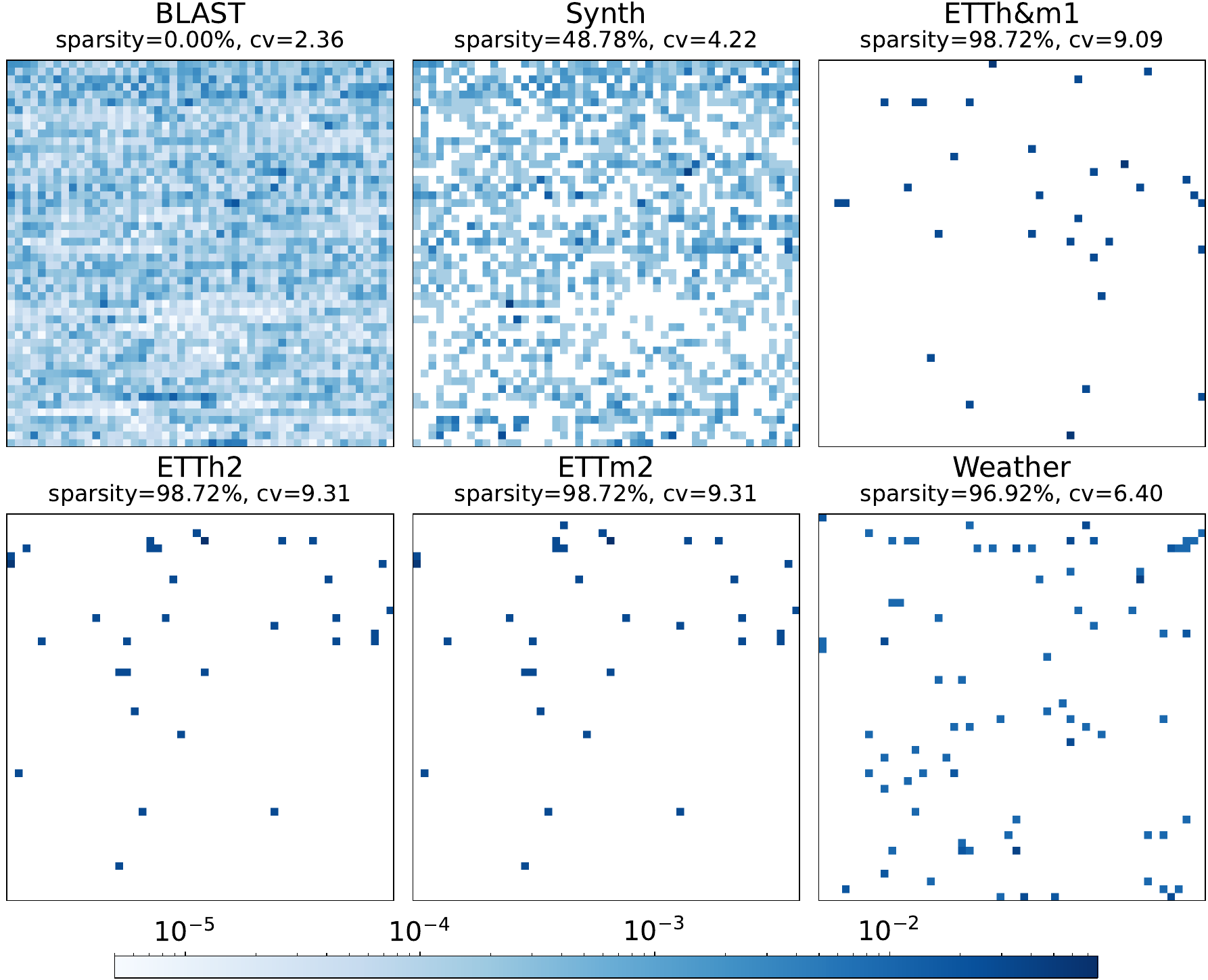}
  \caption{Bin distributions across different datasets.}
  \label{fig:distribution}
\end{figure}

\mypara{Distributional Bias.}
Figure~\ref{fig:distribution} shows large disparities in pattern-bin occupancy: ETT and Weather cover only 2\% to 4\% of the pattern space, while even Synth leaves nearly 50\% unused; BLAST remains concentrated on these benchmark-dominant regions, whereas QUALS enforces more balanced occupancy.

\subsubsection{\textbf{Learnability Analysis}}
This subsection validates {Stage 3} by examining whether the sampling weights learned by learnability synchronization can accelerate the training convergence of universal forecasting models.

\begin{figure*}[htpb]
  \centering
    \setlength{\abovecaptionskip}{0.2cm}
    \setlength{\belowcaptionskip}{0.cm}
  \includegraphics[width=1\linewidth,keepaspectratio]{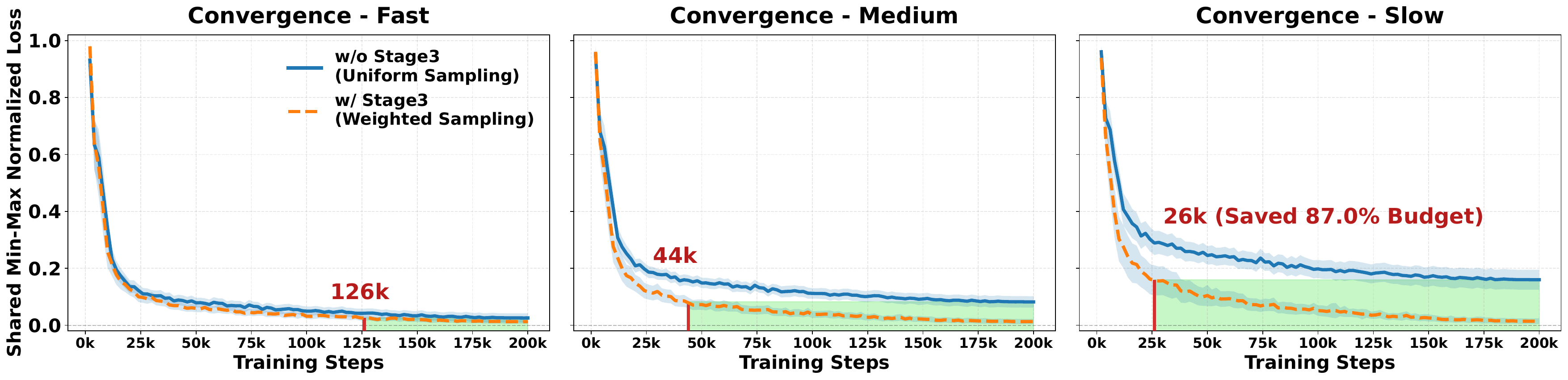}
  \caption{Convergence of \textbf{ChronosBolt}$_{s}$ under uniform vs. weighted sampling across fast, medium, and slow patterns.}
  \label{fig:cconvergence_clustered}
\end{figure*}

Figure~\ref{fig:cconvergence_clustered} visualizes the shared min-max normalized validation loss trajectories of the three groups under uniform and weighted sampling. The fast group shows only a small difference in convergence speed under the two strategies, indicating that uniform sampling already allocates sufficient budget to easy patterns. In contrast, the slow group converges much more slowly under uniform sampling, but is markedly accelerated by weighted sampling. In particular, the weighted strategy reaches the target convergence threshold for the slow group in only 25k steps, saving 87\% of the optimization budget relative to uniform sampling. This improved convergence is also consistent with Table~\ref{tab:performance_efficiency}, where QUALS reduces training tokens while improving the generalization performance of the evaluated models.

\begin{figure*}[htpb]
    \centering
    \newcommand{\subw}{0.33\textwidth}
    \newcommand{\subgap}{\hspace{0.001\textwidth}}
    \begin{subfigure}[t]{\subw}
        \centering
    \setlength{\abovecaptionskip}{0.cm}
    \setlength{\belowcaptionskip}{0.cm}
        \includegraphics[width=\linewidth,page=1]{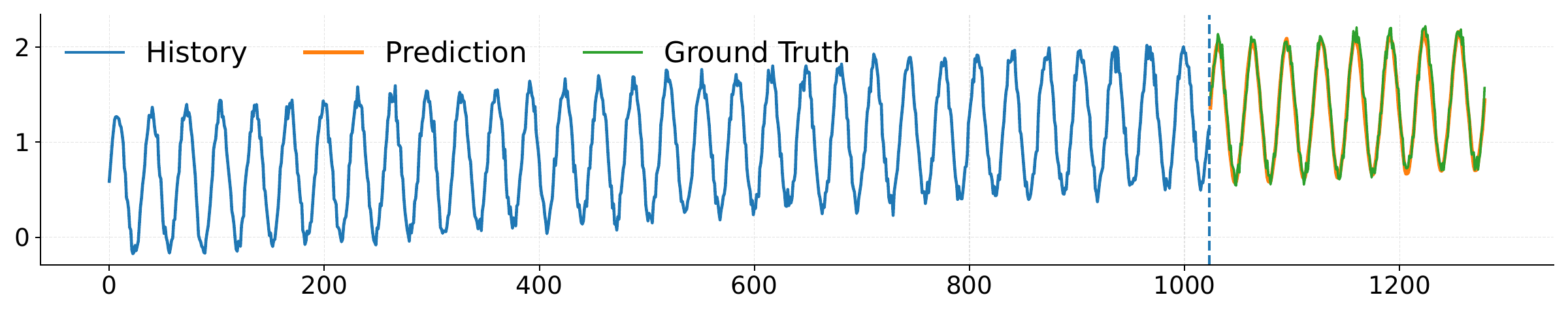}
        \caption{ChronosBolt-Base with QUALS}
        \label{fig:f}
    \end{subfigure}
    \begin{subfigure}[t]{\subw}
        \centering
    \setlength{\abovecaptionskip}{0.cm}
    \setlength{\belowcaptionskip}{0.cm}
        \includegraphics[width=\linewidth,page=1]{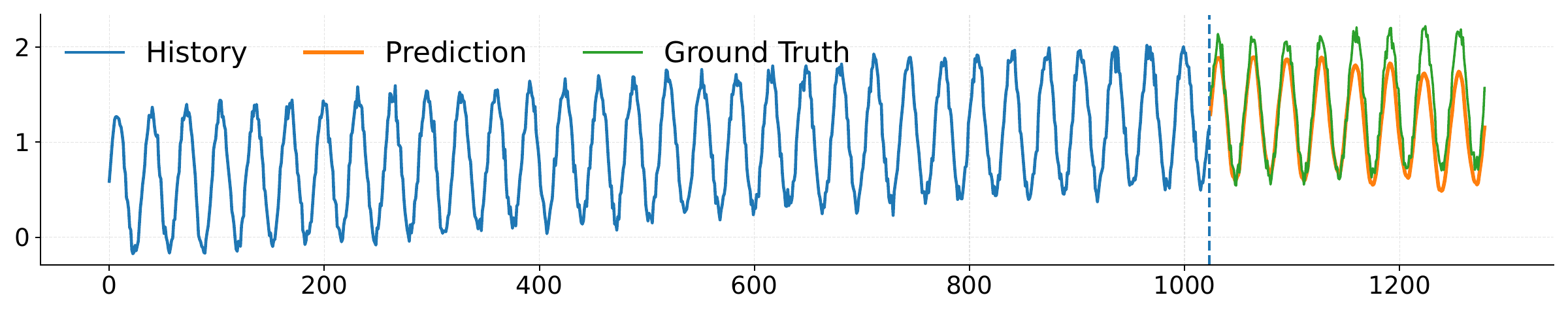}
        \caption{ChronosBolt-Base with BLAST}
        \label{fig:e}
    \end{subfigure}\subgap
    \begin{subfigure}[t]{\subw}
        \centering
    \setlength{\abovecaptionskip}{0.cm}
    \setlength{\belowcaptionskip}{0.cm}
        \includegraphics[width=\linewidth,page=1]{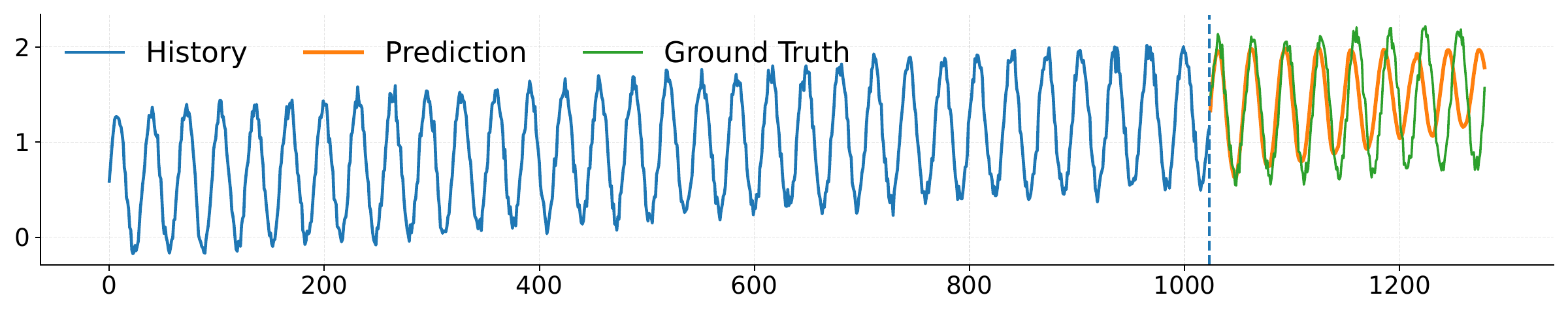}
        \caption{ChronosBolt-Base}
        \label{fig:d}
    \end{subfigure}\subgap

    \begin{subfigure}[t]{\subw}
        \centering
    \setlength{\abovecaptionskip}{0.cm}
    \setlength{\belowcaptionskip}{0.cm}
        \includegraphics[width=\linewidth,page=1]{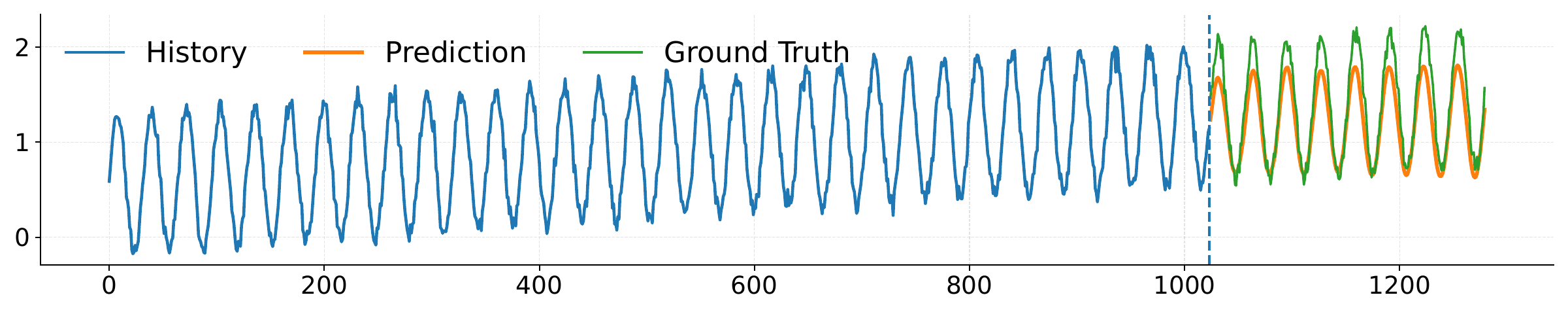}
        \caption{TimeMoE-Large with QUALS}
        \label{fig:l}
    \end{subfigure}
    \begin{subfigure}[t]{\subw}
        \centering
    \setlength{\abovecaptionskip}{0.cm}
    \setlength{\belowcaptionskip}{0.cm}
        \includegraphics[width=\linewidth,page=1]{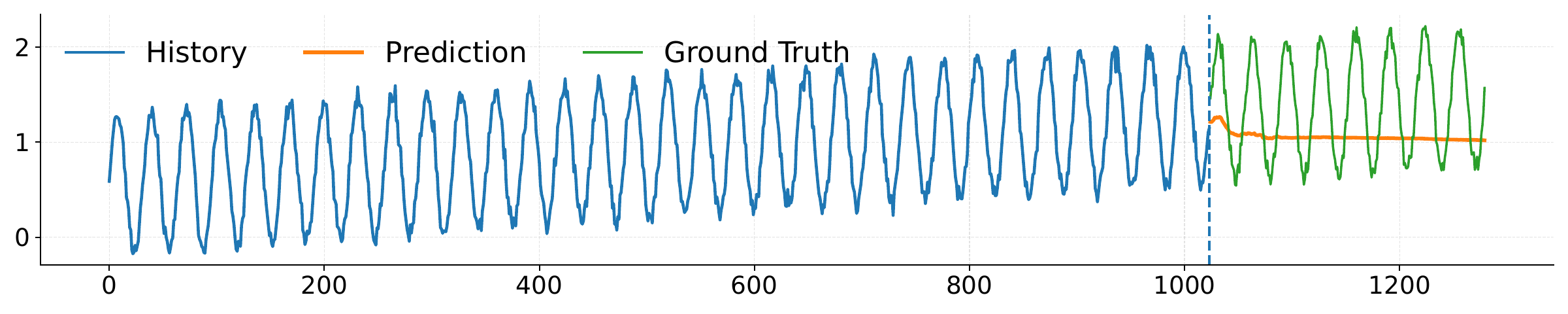}
        \caption{TimeMoE-Large with BLAST}
        \label{fig:k}
    \end{subfigure}\subgap
    \begin{subfigure}[t]{\subw}
        \centering
    \setlength{\abovecaptionskip}{0.cm}
    \setlength{\belowcaptionskip}{0.cm}
        \includegraphics[width=\linewidth,page=1]{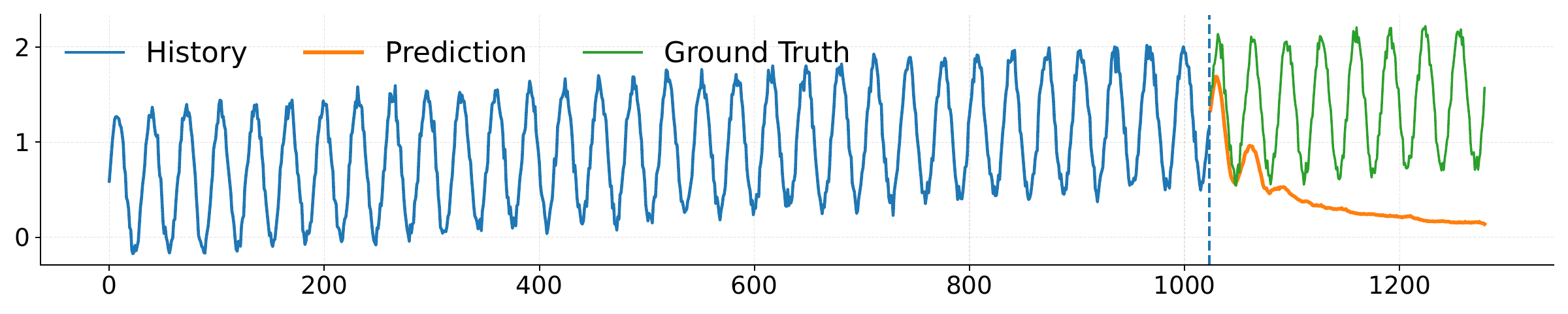}
        \caption{TimeMoE-Large}
        \label{fig:j}
    \end{subfigure}\subgap

    \begin{subfigure}[t]{\subw}
        \centering
    \setlength{\abovecaptionskip}{0.cm}
    \setlength{\belowcaptionskip}{0.cm}
        \includegraphics[width=\linewidth,page=1]{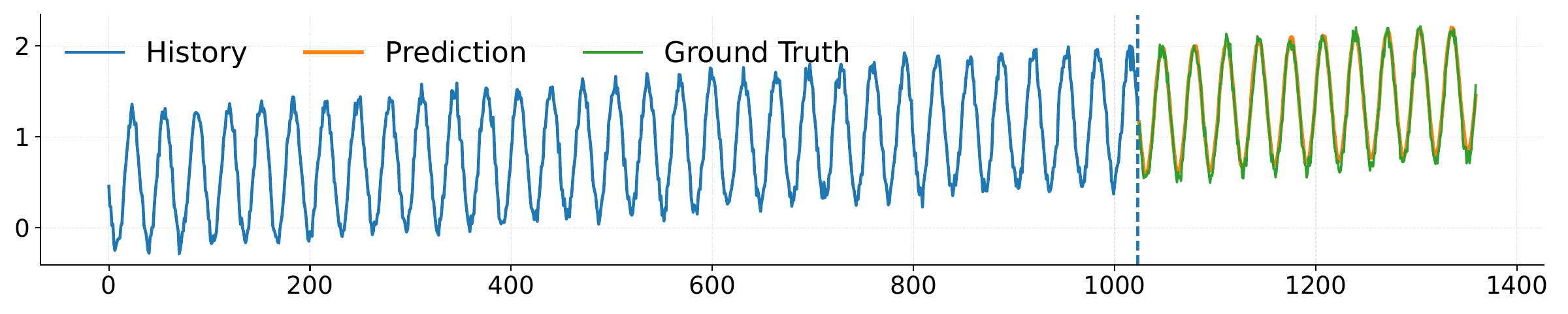}
        \caption{MOIRAI-Large with QUALS}
        \label{fig:r}
    \end{subfigure}
    \begin{subfigure}[t]{\subw}
        \centering
    \setlength{\abovecaptionskip}{0.cm}
    \setlength{\belowcaptionskip}{0.cm}
        \includegraphics[width=\linewidth,page=1]{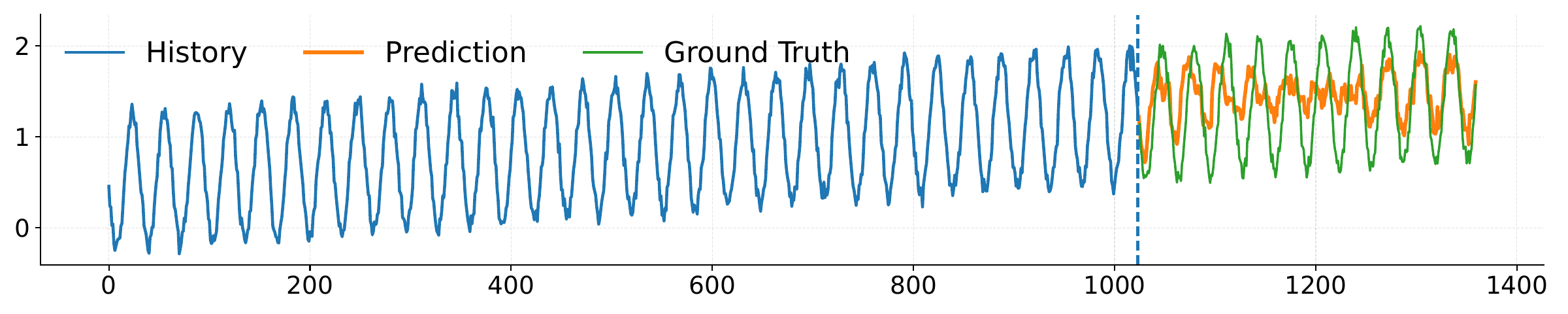}
        \caption{MOIRAI-Large with BLAST}
        \label{fig:q}
    \end{subfigure}\subgap
    \begin{subfigure}[t]{\subw}
        \centering
    \setlength{\abovecaptionskip}{0.cm}
    \setlength{\belowcaptionskip}{0.cm}
        \includegraphics[width=\linewidth,page=1]{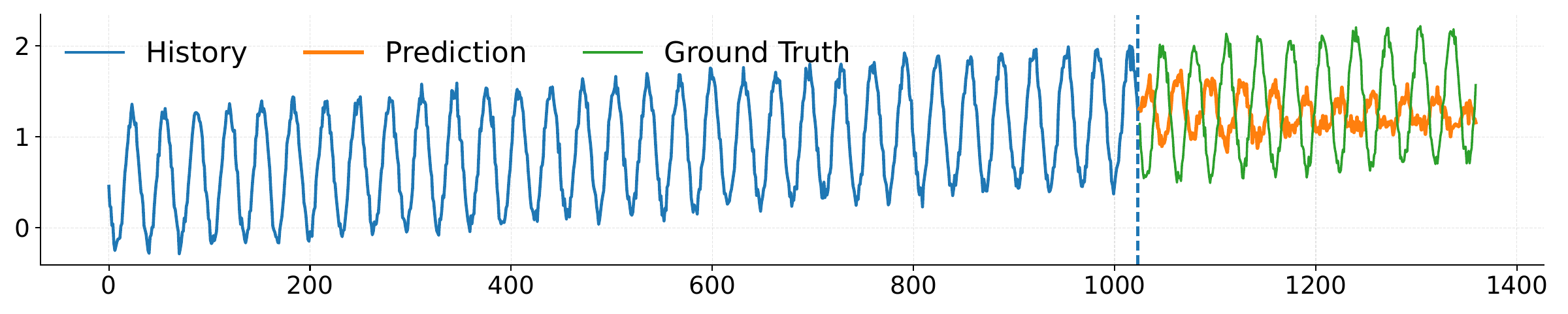}
        \caption{MOIRAI-Large}
        \label{fig:p}
    \end{subfigure}\subgap
    \caption{Comparison of the universal forecasting model trained on the original dataset and on BLAST and QUALS datasets}
    \label{fig:6in2x3}
\end{figure*}

\subsubsection{\textbf{Visualization of Forecast Cases}}
To provide a qualitative view of forecasting behavior, we visualize zero-shot predictions of models pre-trained on the original corpus, BLAST, and QUALS on a synthetic seasonal-trend sequence~\cite{wang2025aries}.

As shown in Figure~\ref{fig:6in2x3}, the original and BLAST-trained models frequently fail to recover the joint structure of trend and seasonality. ChronosBolt mainly exhibits scale mismatch and phase misalignment, TimeMoE is largely unable to model the seasonal component, and MOIRAI often shows distorted seasonal perception. In contrast, QUALS consistently produces more structurally accurate forecasts across all three model families, yielding better recovery of both trend and seasonality.

\section{Conclusion}
\label{sec:conc}


In this paper, we show that unconstrained scaling of pre-training corpora in universal time series forecasting can introduce two important sources of imbalance: skewed pattern distributions and heterogeneous learnability across patterns. To address these issues, we propose QUALS, a three-stage corpus curation framework designed to promote corpus equilibrium. Specifically, QUALS combines VQ-SwinTTS-based pattern quantization, spectral discretization, and learnability synchronization to organize heterogeneous time series into a more balanced pattern space while allocating training emphasis according to optimization difficulty.
Experiments show that models pre-trained on QUALS achieve consistently stronger zero-shot forecasting performance than baseline corpus construction strategies. At the same time, QUALS reduces both data processing overhead and the pre-training token budget required to reach competitive or better performance.
Taken together, these results suggest that corpus curation based on pattern distribution and learnability can provide a more data-efficient alternative to purely scale-driven pre-training for universal time series forecasting.

\clearpage

\balance 

\bibliographystyle{ACM-Reference-Format}
\bibliography{sample}

\end{document}